\documentclass[10pt,twocolumn,letterpaper]{article}
\usepackage{cvpr}

\usepackage{bbm}
\usepackage{duckuments}
\usepackage{multirow}
\usepackage{pifont}
\usepackage{seqsplit}
\usepackage{siunitx}
\usepackage{xfrac}
\usepackage{xspace}
\usepackage{xr} 
\usepackage{algorithm}
\usepackage{algpseudocode}
\usepackage{amsmath}

\definecolor{cvprblue}{rgb}{0.21,0.49,0.74}
\usepackage[pagebackref,breaklinks=true,colorlinks,citecolor=cvprblue,bookmarks=false]{hyperref}

\newcommand{\cmark}{\ding{51}}%

\newcommand{\method}{LagerNVS\xspace}
\newcommand{\Indirect}{Bottleneck\xspace}
\newcommand{\Direct}{Highway\xspace}
\newcommand{\indirect}{bottleneck\xspace}
\newcommand{\direct}{highway\xspace}

\newcommand{\rgt}{\aftergroup\mathclose\aftergroup{\aftergroup}\right}

\newcommand{\xmark}{\ding{55}}%

\newcommand{\image}{I}
\newcommand{\real}{\mathbb{R}}
\newcommand{\camera}{\boldsymbol{g}}

\newcommand{\latent}{\boldsymbol{z}}
\newcommand{\encoder}{e}
\newcommand{\decoder}{h}
\newcommand{\cameraRotation}{\boldsymbol{q}}
\newcommand{\cameraTranslation}{\boldsymbol{t}}
\newcommand{\cameraIntrinsics}{\boldsymbol{k}}
\newcommand{\cameraScale}{\boldsymbol{w}}
\newcommand{\numViews}{V}
\newcommand{\targetCameraTokens}{\boldsymbol{s}}
\newcommand{\loss}{\mathcal{L}}
\newcommand{\numTargetViews}{V_\text{target}}

\makeatletter
\renewcommand{\paragraph}{%
    \@startsection{paragraph}{4}%
    {\z@}{-0.5em}{-0.5em}%
    {\normalfont\normalsize\bfseries}%
}
\makeatother

\def\paperID{6806}
\def\confName{CVPR}
\def\confYear{2026}

\title{
\method:
Latent Geometry for Fully Neural Real-time Novel View Synthesis
}

\author{
Stanislaw Szymanowicz $^{1,2}$
\qquad
Minghao Chen $^{1,2}$
\qquad
Jianyuan Wang $^{1,2}$
\\
Christian Rupprecht $^{1}$
\qquad
Andrea Vedaldi $^{1,2}$
\vspace{2mm}
\\
\centerline{$^1$Visual Geometry Group, University of Oxford \qquad $^2$Meta AI}
}

\begin{document}
\twocolumn[{%
\maketitle
\thispagestyle{empty}
\vspace{-1.3em}%
\begin{center}
\centering
\captionsetup{type=figure}
\vspace{-1em}%
\includegraphics[width=\linewidth]{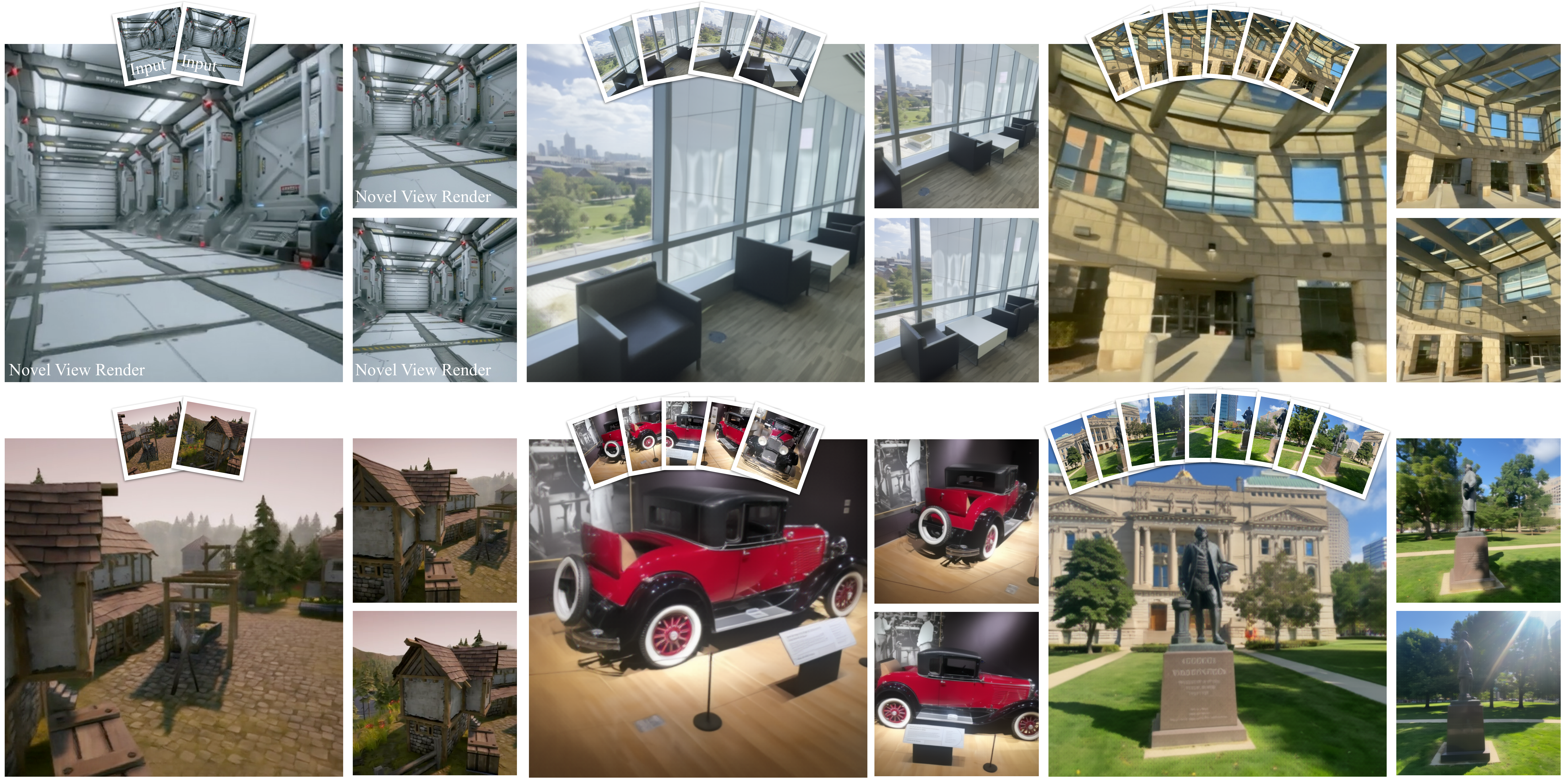}
\captionof{figure}{
\textbf{\method} is a real-time, feed-forward, and generalizable Novel View Synthesis method.
It achieves state-of-the-art quality by rendering views directly with a network, without explicit 3D reconstruction, and by incorporating strong 3D-aware features.
These examples were rendered at more than 30 FPS at 512$\times$512 on a single H100 GPU\@.
}%
\label{fig:teaser}%
\end{center}%

\vspace{0.8em}%
}]

\begin{abstract}
Recent work has shown that neural networks can perform 3D tasks such as Novel View Synthesis (NVS) without explicit 3D reconstruction.
Even so, we argue that strong 3D inductive biases are still helpful in the design of such networks.
We show this point by introducing \method, an encoder-decoder neural network for NVS that builds on `3D-aware' latent features.
The encoder is initialized from a 3D reconstruction network pre-trained using explicit 3D supervision.
This is paired with a lightweight decoder, and trained end-to-end with photometric losses.
\method achieves state-of-the-art deterministic feed-forward Novel View Synthesis (including 31.4 PSNR on Re10k), with and without known cameras, renders in real time, generalizes to in-the-wild data, and can be paired with a diffusion decoder for generative extrapolation.
See 
\href{https://szymanowiczs.github.io/lagernvs}{szymanowiczs.github.io/lagernvs} 
for code, models, and examples.
\end{abstract}
\vspace{-1em}%
\section{Introduction}%
\label{sec:intro}

Novel View Synthesis (NVS) is the task of rendering new views of a scene from a set of other views.
The most common approach to NVS is to fit a 3D model of the scene to the given views via optimization; then, the resulting \emph{explicit 3D reconstruction} is rendered from the target viewpoints~\cite{mildenhall20nerf:,kerbl233d-gaussian}.
This process can work well, but is slow and prone to overfitting unless the number of source views is large.
A recent alternative is to learn a neural network that performs the 3D reconstruction in a feed-forward (optimization-free) manner.
This is faster and can work well even with few or single source views because of the priors learned by the network~\cite{szymanowicz24splatter, charatan24pixelsplat:, chen24mvsplat:, chen2025mvsplat360, smart24splatt3r:, ye2025noposplat, szymanowicz25flash3d, imtiaz25lvt:, xiang25gaussianroom:, jiang25anysplat:}.
The logical next step is to bypass the 3D reconstruction altogether: methods like SRT~\cite{sajjadi21scene}, LVSM~\cite{jin25lvsm:}, and RayZer~\cite{jiang25rayzer:} have shown that the network can output the new views directly.

\begin{figure}
\centering
\includegraphics[width=0.99\columnwidth]{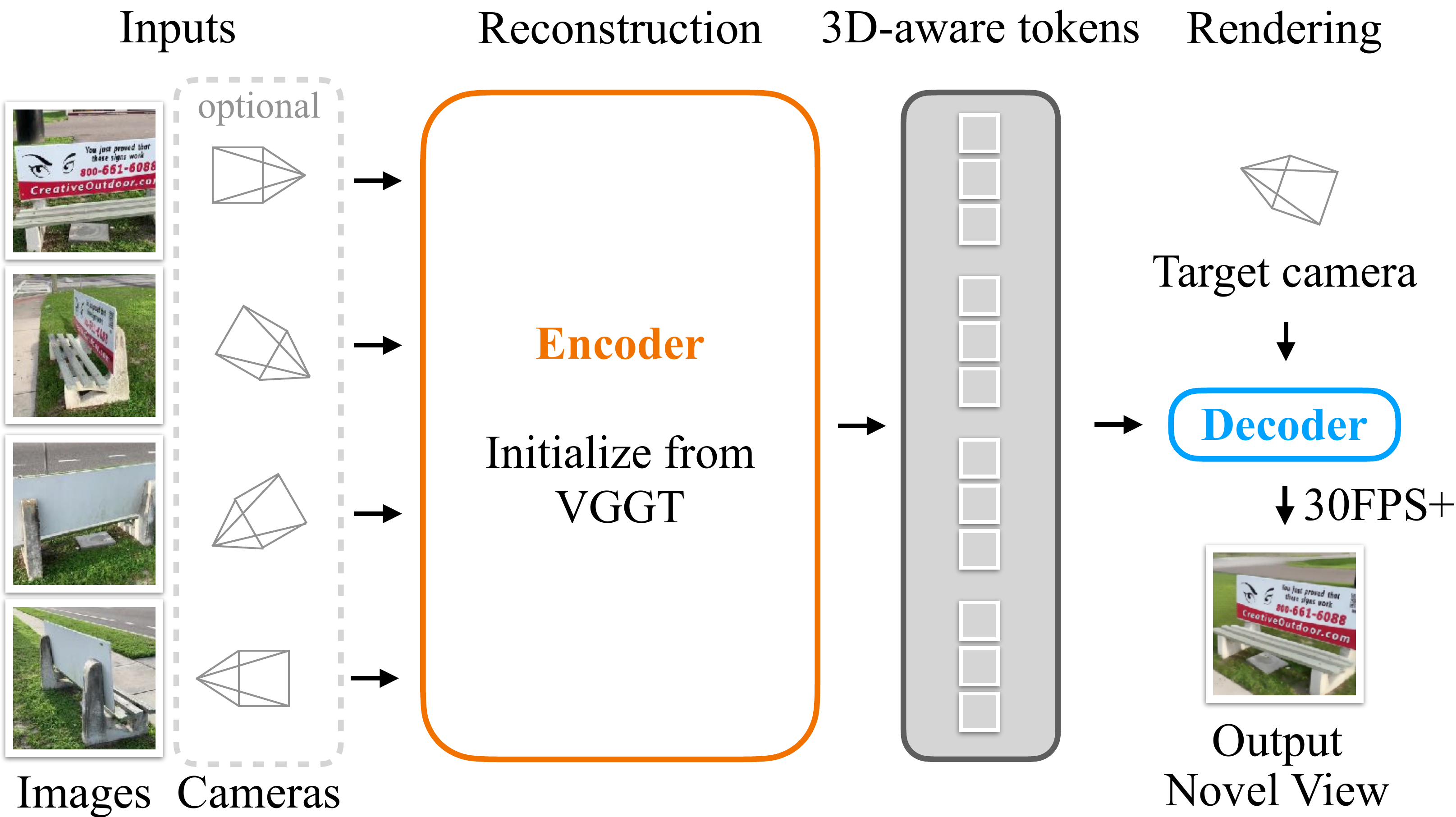}
\caption{\textbf{Method.} The model takes any number of images and, optionally, their camera parameters as input.
A large network initialized from a reconstruction model~\cite{wang2025vggt} outputs an intermediate feature representation with implicit 3D information.
A lightweight network is queried on a target camera pose and renders a $512\times512$ image at more than 30 FPS on a single H100 GPU when provided with up to 9 input images.}%
\label{fig:method}
\end{figure}

However, foregoing 3D reconstruction does not mean that other 3D inductive biases are unimportant.
We show this point by building a NVS network that, like SRT and LVSM, bypasses explicit 3D reconstruction and directly renders the new views.
However, we incorporate 3D-aware features into it (\cref{fig:method}) by initializing the model with the weights of the VGGT~\cite{wang2025vggt} backbone.
This way, we extract features that, while not explicitly `3D', were pre-trained using explicit 3D supervision.
We show that, compared to using strong but generic features like DinoV2~\cite{oquab24dinov2:}, using such 3D-aware features is highly beneficial for NVS\@.

In general, feed-forward NVS architectures are relatively unexplored.
We thus compare several possible designs (\cref{fig:arch_options}).
The simplest is a so-called \emph{decoder-only} architecture that takes the source views and the target camera, and outputs the target views~\cite{jin25lvsm:}.
This works well, but re-evaluates the entire network for each new view generated.
In contrast, \emph{encoder-decoder} architectures~\cite{sajjadi2022srt,jin25lvsm:,jiang25rayzer:} separate encoding from viewpoint-conditioned decoding.
The encoder runs only \emph{once per scene}, extracting a latent 3D representation of it, and only the decoder runs \emph{for each new view}, which can be much more efficient.
We further distinguish `\indirect' encoder-decoders, where the latent tokens are constrained to a fixed dimension before being decoded~\cite{jin25lvsm:}, and `\direct' encoder-decoders, where the decoder can access all image features directly.
We show that the latter strikes an excellent quality and speed trade-off.

Based on these insights, we build \emph{\method}, a \textbf{La}tent \textbf{Ge}ometry model for \textbf{R}eal-time NVS (\cref{fig:teaser}).
\method performs significantly better than LVSM, ($+1.7$dB PSNR margin on the standard RealEstate10k~\cite{zhou18stereo} benchmark), the previous state-of-the-art in reconstruction-free NVS\@.
It also outperforms feed-forward 3D reconstruction networks, including those~\cite{jiang25anysplat:} that are based on VGGT\@.

We also show that training the network on a large mixture of datasets is important for NVS quality and generalization.
Compared to prior models that are usually trained on single datasets, \method works well on ego-centric, 360$^\circ$, and non-square images, as well as images collected in-the-wild, even when source camera poses are unknown, with a single set of model parameters (\cref{fig:generalization}).

\method is also efficient: encoding requires mere seconds and decoding runs in real time (30FPS$+$) with up to nine source images at 512$\times$512 resolution on a single H100 GPU\@.
This is notable because the renderer/decoder is a standard neural network which does not use explicit 3D representations, custom kernels~\cite{kerbl3Dgaussians} or JIT compilation~\cite{ansel2024pytorch2}.

Like its peers, \method is trained using a regression loss, which tends to regress to the mean in the presence of ambiguity, such as when rendering parts of the scene that are not visible in the source views.
This calls for \emph{generative} models that can sample \emph{plausible} views when information is missing.
Motivated by this, we repurpose \method's decoder for denoising diffusion~\cite{ho20denoising}, with promising results.

To summarize, our contributions are as follows:

\begin{figure*}[t]
\centering
\includegraphics[width=0.99\textwidth]{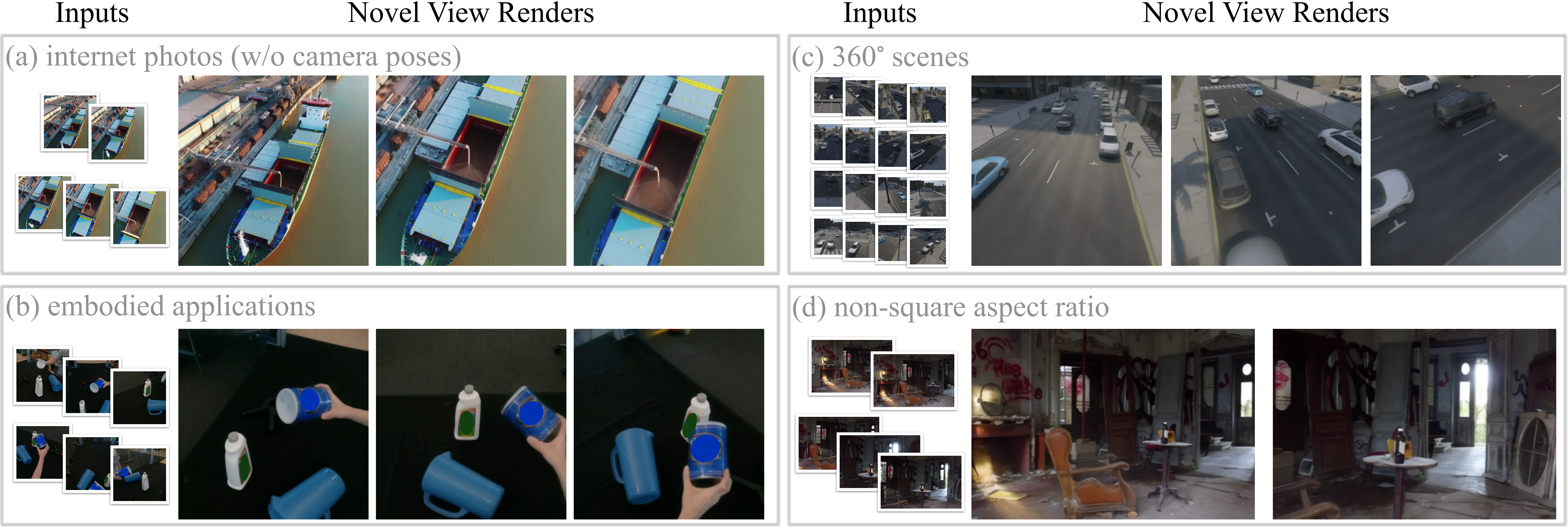}
\caption{\textbf{Generalization.} Our model generalizes to a wide range of input types and applications, including (a) internet photos with unknown camera poses, (b) embodied applications (logos removed as postprocessing), (c) 360$^{\circ}$ scenes and (d) non-square aspect ratios.}%
\label{fig:generalization}
\end{figure*}

\begin{enumerate}
\item We show that reconstruction-free NVS still benefits from strong 3D biases implicitly captured by features pre-trained using explicit 3D supervision.

\item We explore three NVS architectures, decoder-only, \indirect encoder-decoder, and \direct encoder-decoder, and show that the latter has the best rendering quality for a given size of the renderer.
Moreover, we propose a decoder design that enables real-time rendering on a single H100 with up to 9 source views at $512\times512$ resolution.

\item We set the new SoTA for deterministic NVS, outperforming the prior SoTA, LVSM, by a solid $+1.7$dB margin.
We also outperform NVS models that \emph{do} perform feed-forward 3D reconstruction, both with and without source camera poses.
\end{enumerate}

We release several model checkpoints and code to reproduce and further extend our results (see the project \href{https://szymanowiczs.github.io/lagernvs}{website}).
A model performance card is given in~\cref{tab:high_res_generalizable}.
\section{Related work}%
\label{s:related}

\paragraph{Encoder-decoder NVS with explicit 3D representations.}

Many NVS methods are based on reconstructing the scene, extracting an \emph{explicit} 3D representation of it.
By explicit, we mean that the representation maps 3D locations to corresponding local properties like opacity and radiance.
Neural Radiance Fields~\cite{mildenhall20nerf:} and 3D Gaussians~\cite{kerbl233d-gaussian} fit (encode) explicit 3D scene representations to the source views via optimization, which is slow and overfits unless hundreds of views are given.
Hence, authors have also proposed neural networks that can extract 3D representations in a feed-forward manner, quickly and from only a few views.
Some output NeRFs~\cite{yu21pixelnerf:,lin23vision,chen21mvsnerf:} and others per-pixel 3D Gaussians~\cite{%
szymanowicz24splatter,%
charatan24pixelsplat:,
chen24mvsplat:,%
szymanowicz25flash3d,%
gslrm2024,%
ziwen2024longlrm,%
xu2025depthsplat,%
ye2026yonosplat}.
Many methods assume known source cameras, but others relax this constraint~\cite{
jiang25anysplat:,%
ye2025noposplat,%
hong2024coponerf,%
smart24splatt3r:,%
zhang25flare:,%
ye2026yonosplat},
often by using pre-trained multi-view reconstruction models~\cite{wang24dust3r:,duisterhof24mast3r-sfm:,wang2025vggt,wang2026pi3}.
We too leverage such models, but use their latent feature representation instead of their explicit 3D outputs.

\paragraph{Encoder-decoder NVS with latent 3D representations.}

Other approaches extract a \emph{latent} 3D representation of the scene that can be decoded directly into new views, but not necessarily into explicit 3D properties.
These representations can be viewed as an encoding of the scene's \emph{light field}~\cite{adelson91the-plenoptic,gortler96the-lumigraph}, a concept associated with NVS~\cite{buehler01unstructured}.
Early neural approaches like LFN~\cite{sitzmann21light} used auto-decoding to fit compact light field representations.
SRT~\cite{sajjadi2022srt} later proposed an encoder-decoder network to extract such representations in a feed-forward manner from source views.
LVSM~\cite{jin25lvsm:} further improved this approach with increased decoder capacity, while RayZer~\cite{jiang25rayzer:} enabled training without camera labels for ordered image collections.
Like LVSM, we too use a transformer-based architecture, but we (1) propose a different encoder information flow, and (2) leverage a pre-trained 3D reconstruction network for our encoder.
Both (1) and (2) substantially improve performance.
Unlike RayZer, we can operate on unordered image collections.
Concurrently to us, SVSM~\cite{kim2026svsm} analyzed the scaling laws of encoder-decoder NVS transformers to maximize training efficiency.
We use a similar architecture, but instead of optimizing compute usage, we focus on the role of 3D pre-training and on how it enables inference both with and without known cameras, with strong generalization.

\paragraph{Decoder-only NVS.}

Decoder-only methods directly map source images and a target camera to the target image without extracting a camera-independent intermediate representation, thus requiring the entire model to run for every new view rendered and limiting rendering speed.
LVSM~\cite{jin25lvsm:} also considers a variant that follows this paradigm.

\paragraph{Generative NVS.}

NVS is ambiguous when the target camera points at a part of the scene that is not represented in the source images.
In such a case, a \emph{generative} model that can hallucinate \emph{plausible} completions is required.
Generative diffusion decoders were used both in decoder-only NVS methods~\cite{watson23novel,gao24cat3d:,wu25cat4d:,jensen25stable,sargent23zeronvs:,hoorick24generative,liu23zero-1-to-3:,bai25recammaster:,bahmani25lyra:,szymanowicz25bolt3d:,liang24wonderland:},
and in encoder-decoder methods~\cite{wu2024reconfusion,chan23generative,ren25gen3c:,gu23nerfdiff:,chen2025mvsplat360,fischer25flowr:,liu24reconx:, yu24viewcrafter:}.
While we focus on deterministic NVS, we include a preliminary experiment to demonstrate how our model can support diffusion-based generation too.
\section{Method}%
\label{s:method}

In \emph{Novel View Synthesis} (NVS), we are given $\numViews$ source images
$
\image_1, \dots, \image_\numViews \in \real^{3\times H\times W}
$
and the parameters $\camera$ of a target camera, expressed with respect to the viewpoint of image $I_1$ taken as \emph{reference}.
The goal is to output a new image $\image \in \real^{3\times H \times W}$ captured by the target camera $\camera$, which we write as a function
$
\image = f(\camera; \image_1, \dots, \image_\numViews).
$
If the camera parameters $\camera_1, \dots, \camera_\numViews$ of the source images are also known, then the function becomes
$
\image = f(\camera; \image_1, \camera_1, \dots, \image_\numViews, \camera_\numViews)
$.

In practice, we further assume that the focal length is the same across all source and target cameras, and that the vertical and horizontal focal lengths are equal.
When the input cameras are unknown, we pass to the model a target camera with the horizontal field of view set to the canonical value $k_0$, but still train it to render an image at the (unknown) source focal length.%
\footnote{
In the first version of the paper, we did not make this distinction and effectively ``leaked'' the source focal length to the model that is not supposed to receive source cameras as input.
Using this canonical value for the target focal length fixes this corner case with marginal impact on the results.
See the appendix and limitations for details.
}

Following~\cite{wang2025vggt}, we parameterize the cameras with 11-dimensional vectors as
$
\camera = (\cameraRotation, \cameraTranslation, \cameraIntrinsics, \cameraScale)
$,
where
$\cameraRotation \in \mathbb{S}^3$,
is the camera rotation (expressed as a unit quaternion),
$\cameraTranslation \in \real^3$,
is the camera translation,
$\cameraIntrinsics \in \real_+^2$ are the horizontal and vertical fields-of-view (the optical center is assumed to be at the image center).
All $\camera_i$ are relative to the first camera $\camera_1$.
We also introduce an auxiliary input parameter $\cameraScale \in \real_+^2$ to represent the scene scale, discussed in the supplement.

When $f$ is implemented as a neural network, we call the model \emph{decoder only} if the whole network is evaluated for each new $\camera$.
We call it an \emph{encoder-decoder} if it first encodes the source images into an intermediate representation $\latent$ independently of the target camera, so that the NVS function $f$ is the composition of an encoder $\encoder$ and a decoder $\decoder$:
\begin{equation}
\latent = \encoder(\image_1, \camera_1, \dots, \image_\numViews, \camera_\numViews),
\quad
\image = \decoder(\latent; \camera).
\end{equation}
This allows us to amortize the cost of computing $\latent$ across the generation of multiple target views.

\begin{figure}[t]
\centering
\includegraphics[width=\columnwidth]{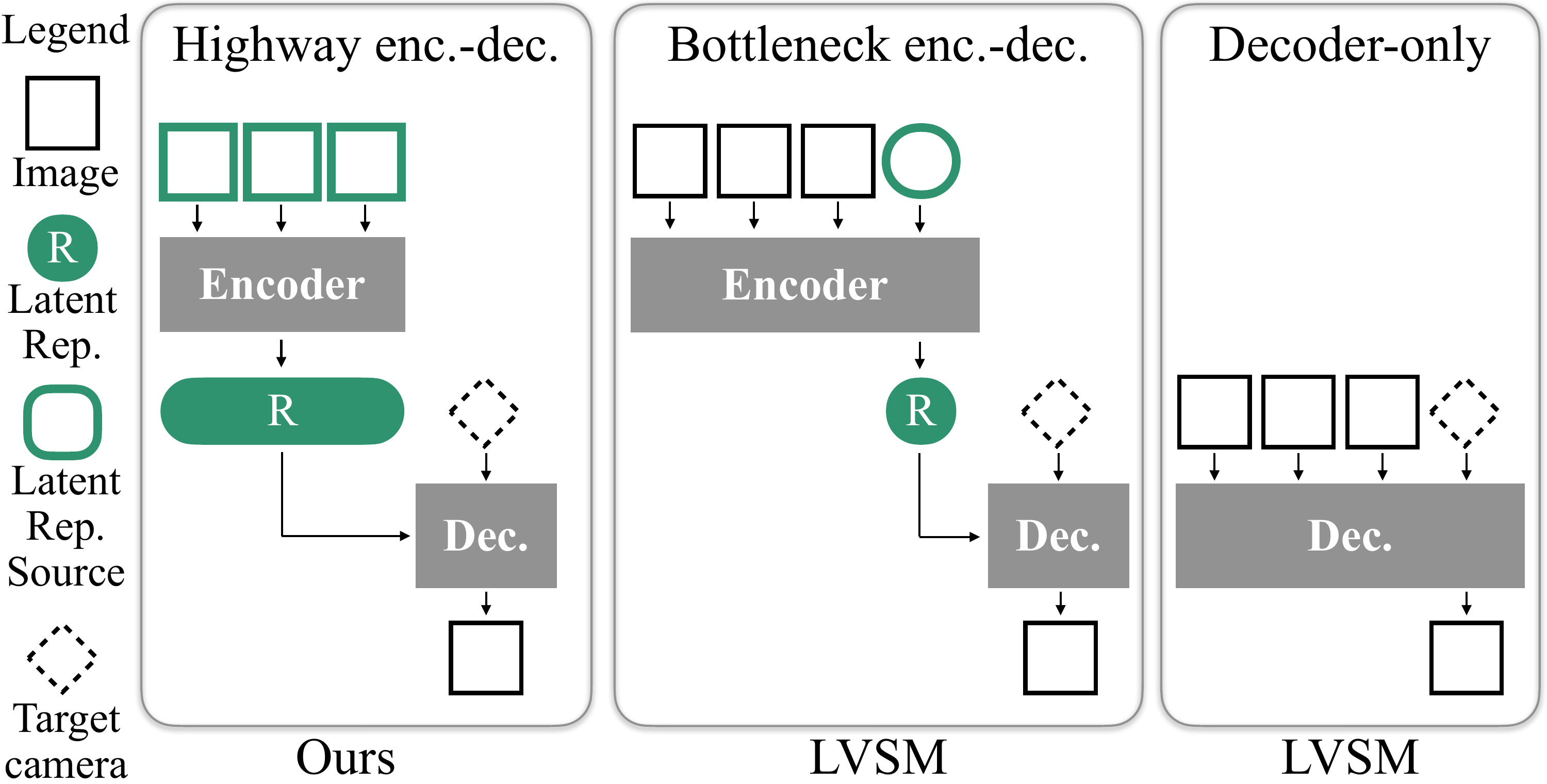}
\caption{\textbf{Architectures.}
Our \direct encoder-decoder (left) maximizes information flow from source images to the latent representation (R), making it more expressive than the \indirect encoder-decoder (middle).
Unlike decoder-only models (right), our design allows scaling the encoder without slowing decoding.
}%
\label{fig:arch_options}
\end{figure}

We further distinguish ``\direct'' and ``\indirect'' encoder-decoders (\cref{fig:arch_options}).
In \emph{\direct} encoder-decoders (named so to reflect non-attenuated information flow~\cite{srivastava2015highway}), the representation $\latent = (\latent_1,\dots,\latent_\numViews)$ contains separate feature vectors for each source image, whereas in \emph{\indirect} ones the number of tokens in $\latent$ is independent of the number $\numViews$ of source views, thus constraining information flow.

While we are particularly interested in models where both the encoder $\encoder$ and the decoder $\decoder$ are neural networks, encoder-decoders can also be based on explicit 3D reconstruction.
In this case, $\latent$ is a 3D representation, such as 3D Gaussians, and $\decoder$ is the corresponding renderer.

\subsection{An encoder with an implicit 3D bias}%
\label{sec:inductive-biases}

One obvious way to incorporate 3D inductive biases in the model $f$ is to opt for a reconstruction approach where $\latent$ is an explicit 3D representation of the scene.
A recent example is AnySplat~\cite{jiang25anysplat:}, which builds on VGGT~\cite{wang2025vggt} for its encoder and uses a Gaussian splat renderer for the decoder.
Alternatively, both the encoder and decoder can be neural networks, and the representation $\latent$ can be a set of features~\cite{sajjadi2022srt,jin25lvsm:,jiang25rayzer:}, avoiding explicit 3D reconstruction and representations altogether.
In this case, there are few 3D inductive biases, mostly limited to choosing an encoder for the camera parameters (e.g., ray maps).

Here we explore a third approach (\cref{fig:method}), where encoder and decoder are neural networks, but the encoder is initialized from a network pre-trained for 3D reconstruction via explicit 3D supervision.
In this way, there is no explicit 3D reconstruction, but the features $\latent$ output from the encoder are ``3D-aware''.

We call our model \method, and we implement its encoder by building on top of the VGGT model~\cite{wang2025vggt}.
Recall that VGGT is a feed-forward 3D reconstruction network that maps one or more images of a scene to a number of geometric quantities, including the camera $\camera_i$ and the depth map $D_i$ for each source image $\image_i$.
However, we do not make use of these outputs directly.
Instead, for each source image $\image_i$, we extract an array of tokens
$
\latent_i \in \mathbb{R}^{P \times C}
$
from the last layers of VGGT's transformer backbone (before the decoding heads)---we extract the tokens output by the last local attention layer and the last global attention layer, and remove the so-called camera tokens.
We concatenate these two sets of tokens channel-wise, pass them through a linear layer to project them to the desired dimension $C$ that our decoder expects, and normalize with LayerNorm~\cite{ba16layer} to improve learning stability.

\paragraph{Augmenting the encoder with camera inputs.}

In applications where the source cameras $\camera_1, \dots, \camera_\numViews$ are available, we wish to pass them to the encoder too.
Since VGGT does not take cameras as input, we modify it by adding a 2-layer Multi-Layer Perceptron (MLP) that projects the camera parameters $\camera$ to a $1024$-dimensional token.
When a camera is not provided as input, we set $\camera$ to the null vector (keeping only the scene scale parameter $\cameraScale$).
We add the projected tokens to the VGGT default initial values for the camera tokens, and then feed these to the VGGT feature backbone as usual.
With these modifications, the encoder $\encoder$ is now a function of images and optional cameras.

\subsection{An efficient decoder}%
\label{sec:decoder}

The decoder $\decoder(\camera;\latent_1,\dots,\latent_\numViews)$ takes as input the tokens $\latent$ output by the encoder (\cref{sec:inductive-biases}) and the target camera $\camera$.

\paragraph{Encoding the target camera.}

The goal of the decoder is to \emph{render} the image from the viewpoint of the target camera $\camera$, and we represent it densely in the form of a \emph{Plucker ray map}~\cite{zhang24cameras}.
Hence, $\camera$ is input to the decoder as a $6 \times H \times W$ image where each pixel contains its corresponding camera ray in Plucker coordinates, namely $(r_d, r_m)$ where $r_d \in \real^3$ is the ray direction and $r_m\in\real^3$ is the ray moment.
When the source camera intrinsics are unknown, we form target camera rays with a nominal horizontal field of view $\cameraIntrinsics_x=\pi / 2$ to prevent information leakage (see above).
A convolutional layer with kernel size and stride $r'=8$ is then applied to this image to extract $HW/{r'}^2$ tokens.
Four additional register tokens~\cite{darcet24vision} are concatenated to these.
We use the symbol $\targetCameraTokens$ to denote the resulting set of $4 + HW/{r'}^2$ tokens encoding the target camera $\camera$.

\paragraph{Decoder architecture.}

We design the decoder as a transformer network, where the dense camera tokens $\targetCameraTokens$ attend the encoded source images $\latent_1, \dots, \latent_\numViews$ to form the target image.
We experiment with two variants (see 
Alg. 1
in the supplement) trading off quality and speed.
The first variant has complexity $\mathcal{O}(V^2)$ and uses \emph{full attention} on the concatenation of target camera and scene tokens $(\targetCameraTokens, \latent_1, \dots, \latent_\numViews)$, using them as queries, keys and values in the transformer:
\begin{equation}\label{eq:full_attn}
    q=k=v =(\targetCameraTokens, \latent_1, \dots, \latent_\numViews).
\end{equation}
The second variant has complexity $\mathcal{O}(V)$ and uses full attention on the target camera tokens $\targetCameraTokens$ only, setting $q = k = v = \targetCameraTokens$, followed by \emph{cross attention} between these and the scene tokens, with two layers that use respectively
\begin{equation}\label{eq:cross_attn}
\begin{split}
    q_1&=\targetCameraTokens; \quad k_1 = v_1= (\latent_1, \dots, \latent_\numViews), ~~\text{and} \\
    q_2&=( \latent_1, \dots, \latent_\numViews ); \quad k_2 = v_2= \targetCameraTokens
\end{split}
\end{equation}
We otherwise use a standard transformer architecture (see \cref{sec:details} and the supplement).
At the output, the register tokens are discarded, and the dense target camera tokens are projected with a linear layer to $8\times8$ patches and reshaped to the original image size to obtain the target image $\image$.

\subsection{Training}%
\label{sec:training}

We train our model by minimizing the NVS loss, i.e., the distance between ground truth novel views and the ones estimated by the model.
In particular, we use a combination of mean-squared (L2) error and perceptual~\cite{johnson16perceptual} losses, i.e.,
$
\loss =
\lambda_{2} \loss_2 +
\lambda_{p} \loss_p
$.

Given that we start from a pre-trained VGGT model for our encoder, we have a choice on whether to fine-tune the entire model end-to-end, or restrict learning to the new parameters, most of which reside in the decoder.
Empirically, we found that fine-tuning the entire model is essential to obtain good results.
This is perhaps not surprising as the VGGT features were not trained with the goal of retaining the \emph{appearance} of the source images or with the capability to understand camera pose conditioning.

\paragraph{Data.}

To train our model, we thus require a collection of tuples
$
(\image, \camera, \image_1, \camera_1, \dots, \image_\numViews, \camera_\numViews)
$,
consisting of posed images of approximately static 3D scenes.
Inspired by VGGT, we train on a rich mix of 13 multi-view datasets (see supplement), including typical NVS datasets such as
RealEstate10k~\cite{zhou18stereo},
DL3DV~\cite{ling23dl3dv-10k:},
WildRGBD~\cite{xia24rgbd}.
Our full dataset roughly matches the size and diversity of the data used for VGGT\@.

\subsection{Implementation details}%
\label{sec:details}

\paragraph{Architecture.}

We use the pre-trained VGGT model for our encoder and start from its pre-trained weights.
We use a ViT-B~\cite{dosovitskiy21an-image} transformer for our decoder with FlashAttention~\cite{dao2022flashattention,dao2023flashattention2,shah24flashattention-3:} attention kernels.

\paragraph{Augmentations.}

We train our model to be robust to the inputs to the model: we randomly sample the number of source views, varying between 1 and 10, selectively drop out camera tokens and vary the aspect ratio.

\paragraph{Optimization.}

We use AdamW optimizer and cosine learning rate decay with linear warmup.
We use QK-norm~\cite{henry20query-key}, and gradient clipping for improved training stability, as well as gradient checkpointing for reduced memory usage.
Our main model is trained at $512$ resolution (longer side) on the full dataset mix for $250$k iterations with batch size 512, but we adjust hyperparameters (e.g., batch size, learning rate and training iterations) to the dataset and baselines used in each experiment.

\section{Experiments}%
\label{s:experiments}

We begin by comparing \method to LVSM, the prior SoTA for NVS (\cref{sec:exp-sota}), and by demonstrating the benefits of training the model on a large number of different datasets (\cref{sec:exp-generalization}).
Then, we study the importance of 3D-aware pre-training and the choice of encoder-decoder architecture (\cref{sec:exp-ablations}).
We also show that our method, which uses an implicit 3D representation, outperforms methods that reconstruct the 3D scene explicitly (\cref{sec:exp-explicit}).
Finally, we demonstrate how our model can be used in combination with diffusion models (\cref{sec:exp-diffusion}).

We assess \method in terms of its ability to deliver high-quality NVS for a given complexity of the decoder, as the latter ultimately determines the rendering speed, which we aim to maintain in the real-time range.
In all comparisons we thus treat the number of decoder transformer blocks as a control variable.
We describe the most important experimental parameters and results here, and refer the reader to the supplement for details.

\paragraph{Evaluation datasets.}

We test our method on 3 datasets commonly used for Novel View Synthesis: RealEstate10k~\cite{zhou18stereo}, DL3DV~\cite{ling23dl3dv-10k:}, and CO3D~\cite{reizenstein21co3d}.
We adapt the training and testing setup to match that of our baselines, and include more details in each sub-section.

\paragraph{Metrics.}

On all quantitative tasks we use standard metrics: Peak Signal-to-Noise Ratio (PSNR), Structural Image Similarity (SSIM~\cite{wang04bimage}), and Perceptual Similarity (LPIPS~\cite{zhang18the-unreasonable}).

\subsection{Comparison to the state-of-the-art in NVS}%
\label{sec:exp-sota}

\begin{figure}
\centering
\includegraphics[width=\columnwidth]{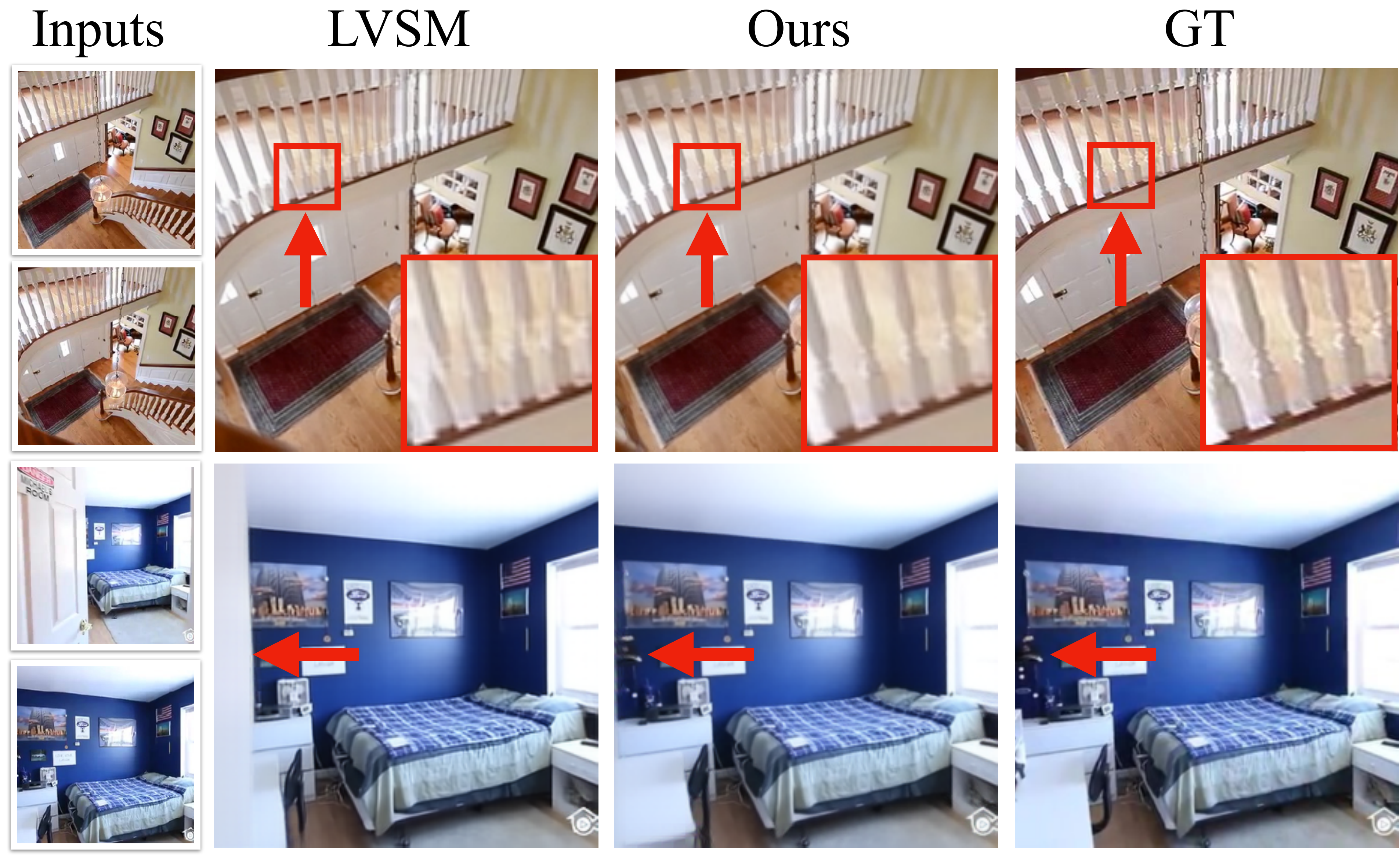}
\caption{\textbf{Comparison to LVSM.}
Our model estimates geometry better than LVSM in regions where 2D matching is challenging (top).
It also uses monocular cues better for depth (bottom).}%
\label{fig:vs_lvsm_re10k}
\end{figure}

\begin{table}
\centering
\footnotesize
\resizebox{0.97\columnwidth}{!}{%
\begin{tabular}{lllcccc}
\toprule
       & Method                                    & Batch      & PSNR $\uparrow$ &  SSIM $\uparrow$ & LPIPS $\downarrow$ \\
\midrule
(a)    & LVSM~\cite{jin25lvsm:} Enc-dec.~\indirect & 64         & 28.32           &  0.888           & 0.117              \\
(b)    & LVSM~\cite{jin25lvsm:} Decoder-only       & 64         & 28.89           &  0.894           & 0.108              \\
(c)    & Ours Enc-dec.~\direct --- x-attn.         & 64         & 30.11           &  0.912           & 0.089              \\
(d)    & Ours Enc-dec.~\direct --- full            & 64         & \textbf{30.48}  &  \textbf{0.918}  & \textbf{0.086}     \\
\midrule
(e)    & LVSM~\cite{jin25lvsm:} Enc-dec.~\indirect & 512        & 28.58           &  0.893           & 0.114              \\
(f)    & LVSM~\cite{jin25lvsm:} Decoder-only       & 512        & 29.67           &  0.906           & 0.098              \\
(g)    & Ours Enc-dec.~\direct --- x-attn.         & 512        & 31.06           &  0.924           & 0.080              \\
(h)    & Ours Enc-dec.~\direct --- full            & 512        & \textbf{31.39}  &  \textbf{0.928}  & \textbf{0.078}     \\
\bottomrule
\end{tabular}
}
\caption{\textbf{Comparison to LVSM.}
Our model outperforms LVSM (SoTA) both at small (batch size 64) and large (batch size 512) training scales.}%
\label{tab:lvsm_comparison}
\end{table}

We first evaluate the ability of \method to deliver SoTA NVS (\cref{fig:vs_lvsm_re10k,tab:lvsm_comparison}).
The SoTA is LVSM, which, like us, uses a latent 3D representation\@.
We follow LVSM's evaluation protocol and train and test on RealEstate10k~\cite{zhou18stereo} with $V=2$ source views.
The data contains $65$k video clips of indoor scenes.
We use the same $256\times256$ resolution, training steps (100,000), and batch size as LVSM, as well as the source and target views in pixelSplat~\cite{charatan24pixelsplat:}.

\paragraph{Results.}

We test the LVSM \indirect encoder-decoder (\cref{tab:lvsm_comparison} (a), (e)) and decoder-only (\cref{tab:lvsm_comparison} (b), (f)) variants.
The LVSM authors trained their model using batch size 512 (\cref{tab:lvsm_comparison} (e-h)) and 64 (\cref{tab:lvsm_comparison} (a-d)) to simulate resource-constrained training.
We report scores for both settings, and with the full and cross-attention variants of our model.

The LVSM authors noted that their decoder-only models are better than their bottleneck encoder-decoder ones, which they attribute to the model architecture.
However, encoder-decoder architectures can amortize the encoding cost, thus benefiting from more lightweight decoders and consequently faster rendering.
It is thus notable that our \emph{\direct} encoder-decoder is significantly better than both LVSM variants in both settings (up to $+1.7$dB PSNR\@; \cref{tab:lvsm_comparison} (c, d), (f, g)).
\Cref{fig:vs_lvsm_re10k} indicates that our model benefits from improved multi-view matching and monocular depth estimation.
As we further analyze in~\cref{sec:exp-ablations}, this is due to a combination of factors, including using pre-trained 3D-aware features and not having an encoding bottleneck.

\subsection{Generalizable NVS}%
\label{sec:exp-generalization}

While training on a single dataset like Re10k is common practice in many NVS works~\cite{charatan24pixelsplat:,chen24mvsplat:,gslrm2024,jin25lvsm:}, multi-view geometry models~\cite{wang2025vggt,mast3r,wang24dust3r:} have shown the benefits of training on large collections of data, particularly for zero-shot generalization.
Here, we look into this setup as well.

Our full model, trained as described in \cref{sec:details} on all the datasets of \cref{sec:training}, achieves excellent generalization to in-the-wild data and different scene types, and can work without knowledge of the source camera poses, as shown in \cref{fig:generalization}.
It also supports single-view NVS, as shown in \cref{fig:single_view}.
\Cref{tab:high_res_generalizable} in the supplement provides a detailed quantitative evaluation of this model on a number of standard benchmarks.

\begin{figure}
\centering
\includegraphics[width=\columnwidth]{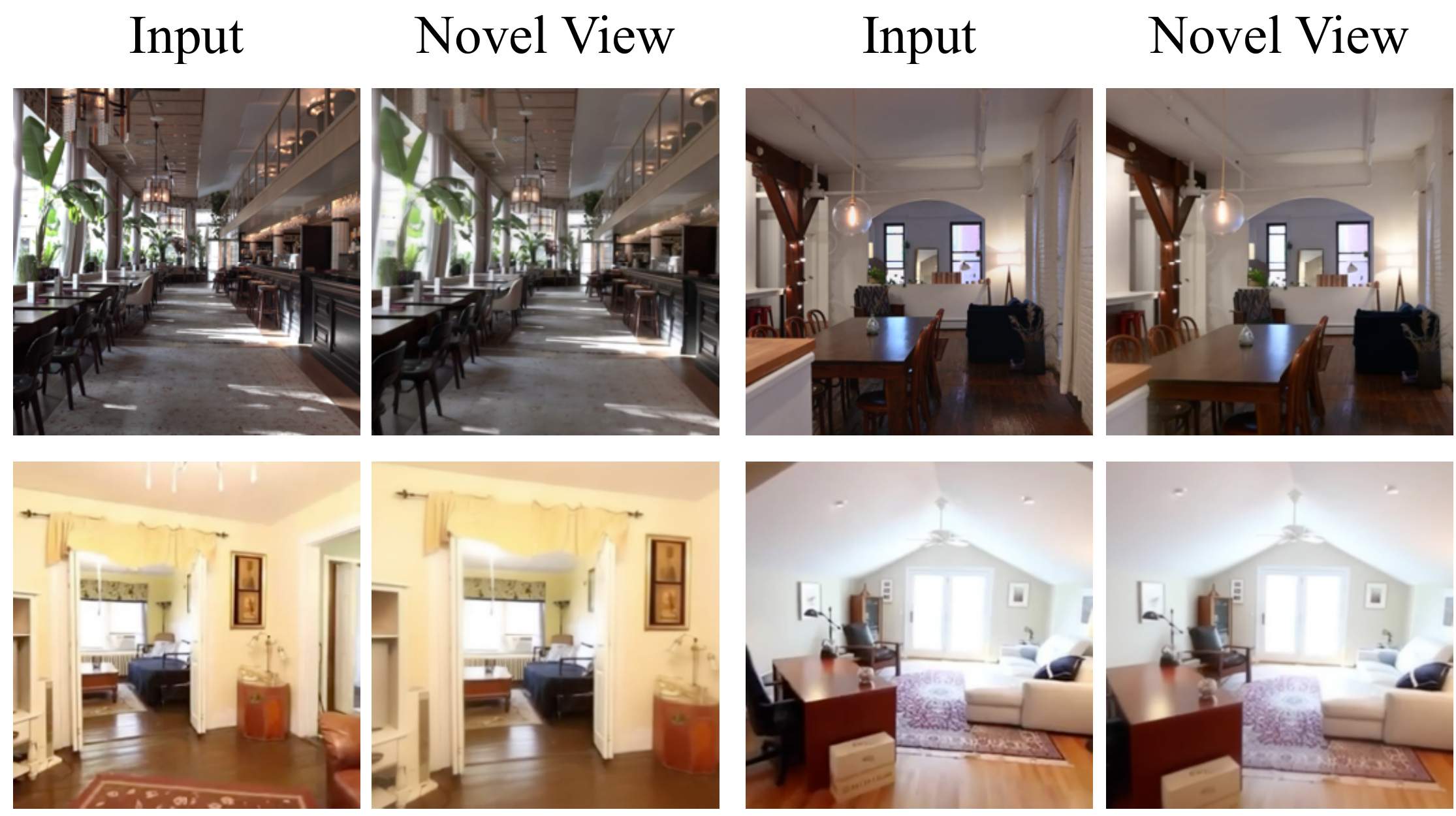}
\caption{\textbf{Single-view NVS} is supported for small camera motion with \method.}%
\label{fig:single_view}
\end{figure}

\subsection{Analysis and ablations}%
\label{sec:exp-ablations}

We assess the importance of 3D pre-training of the encoder features and architectural design choices via ablations.

\paragraph{Experimental protocol.}

As in \cref{sec:exp-generalization}, we train models on all datasets of \cref{sec:training}.
We report results in the 2-view setting with known cameras on the DL3DV split set by DepthSplat with $256\times256$ center-crops, taking care to remove source views from the target view set.
We use the VGGT architecture as the encoder backbone.
We modify the training hyperparameters compared to the main model, including batch size and training iterations, to reduce computation (see supp.).
In the \indirect variant, the global and local aggregator blocks process additional $3072$ bottleneck tokens (following~\cite{jin25lvsm:}).

\paragraph{Importance of using pre-trained 3D-aware features.}

\begin{table}[b]
\centering
\footnotesize
\resizebox{0.97\linewidth}{!}{%
\begin{tabular}{lllllccc}
\toprule
{}                      & Encoder & E2E   & X-attn & Pre-Tr. & PSNR $\uparrow$  &  SSIM $\uparrow$ & LPIPS $\downarrow$ \\
\midrule
\midrule
(a)                   & \Direct  & \cmark & \cmark  & 3D    &  \underline{21.02} & \underline{0.652} & \underline{0.257} \\
(b)                   & \Direct  & \cmark & \xmark  & 3D    &  \textbf{21.30} & \textbf{0.667} & \textbf{0.248} \\
\midrule
(c)                   & Decoder only & \cmark & \xmark  & ---       & 18.39 & 0.533 & 0.407 \\
(d)                   & \Indirect  & \cmark & \xmark  & ---  & 17.53 & 0.480 & 0.461  \\
\midrule
(e)                   & \Direct  & \cmark & \cmark  & ---  & 18.03 & 0.551 & 0.405   \\
(f)                   & \Direct  & \cmark & \cmark  & 2D  & 18.17 & 0.515 & 0.388   \\
(g)                   & \Direct  & \xmark & \cmark  & 3D  & 19.01 & 0.553 & 0.334      \\

\bottomrule
\end{tabular}
}
\caption{\textbf{Ablations.} Even when handicapped by using faster cross-attention, our \direct encoder-decoder model outperforms the decoder-only and \indirect encoder-decoder variants, which are analogous to the models introduced by LVSM~\cite{jin25lvsm:}.
Using 3D-aware pre-training (Pre-tr.) is crucial, but only effective if the features are fine-tuned end-to-end (``E2E'').}%
\label{tab:ablations}
\end{table}

\begin{figure}
\centering
\includegraphics[width=\columnwidth]{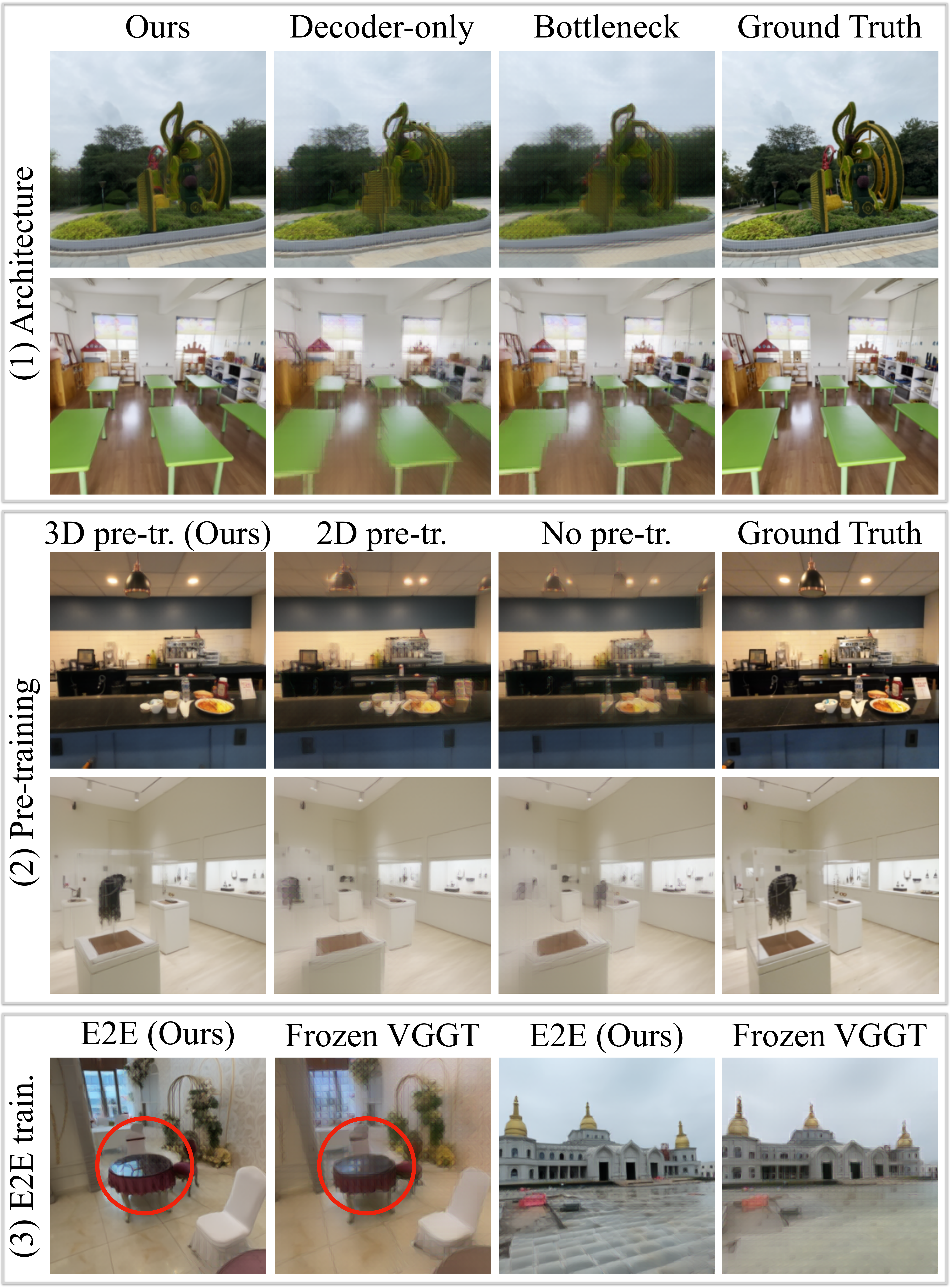}
\vspace{-1.5em}
\caption{\textbf{Ablations.} Our architecture (1), pre-training (2) and end-to-end training (3) result in the best NVS quality.}%
\label{fig:ablations_all}
\vspace{-0.5em}
\end{figure}

\Cref{tab:ablations} shows that initializing the model from VGGT (row (a)) is much better ($+2.9\text{dB}$ PSNR) than starting from scratch (row (e)).
Using generic 2D-pretraining (DinoV2~\cite{oquab24dinov2:} in row (f)) is also marginally better, but not nearly as much as using 3D pre-training.
\Cref{fig:ablations_all} shows that 3D pre-training improves especially the foreground objects, which benefit from better depth estimation.

Another important finding is that it is necessary to fine-tune the entire model end-to-end (E2E) to obtain good performance, unfreezing the VGGT backbone (rows (a) vs (g)).
This is because VGGT is pre-trained to reconstruct only the geometry and thus to treat appearance attributes like colors, reflectance and transparency, which are important for NVS, as nuances.
The results in \cref{fig:ablations_all} corroborate this hypothesis by showing that using a frozen backbone results in a lack of reflections and poor textures.
Freezing VGGT also makes it harder for the model to learn to understand the source cameras (\cref{sec:inductive-biases}) when these are given.

Future versions of VGGT and similar reconstruction models should consider including a rendering head and loss during training to preserve appearance information and thus be more useful for tasks like NVS\@.

\paragraph{Decoder-only vs bottleneck/highway encoder-decoder.}

As in \cref{sec:exp-sota}, the decoder-only model (\cref{tab:ablations} row (c)) outperforms the bottleneck encoder-decoder (row (d)), but our \direct encoder-decoder is better than both (row (a)).
The latter has no bottleneck, can leverage the pre-trained 3D-aware features (VGGT), and, by offloading much of the computation that is independent of the target view to the encoder, has more capacity to learn a high-quality decoder (for a fixed decoding budget, which is our control variable).

The \indirect model's unique advantage is that the decoding speed is independent of the number of source views.
Even so, the cross-attention variant of the \direct architecture still allows real-time rendering (56--30 FPS) with 1--9 source views.

\paragraph{Full vs cross-attention.}

Using full attention (\cref{tab:ablations} row (a)) performs slightly better than cross-attention (row (b)), but is slower: the maximum number of source images for real-time (30FPS+) rendering is $9$ for the cross-attention variant, and $6$ for the full attention variant (see \cref{tab:attn_block} in the supplement), due to the fact that the complexity increases from $O(V)$ to $O(V^2)$ in the number $V$ of source views.

\subsection{NVS with Explicit 3D Representations}%
\label{sec:exp-explicit}

Most NVS methods reconstruct the scene, extracting an explicit 3D representation of it, which also provides a strong 3D inductive bias.
In this section, we show that our \emph{latent} 3D representation performs better than these solutions.

\paragraph{Baselines and evaluation protocols.}

\begin{table}
\centering
\footnotesize
\resizebox{\columnwidth}{!}{%
\begin{tabular}{lllccc}
\toprule
                          & Method                                             & Test data                 & PSNR $\uparrow$         & SSIM $\uparrow$         & LPIPS $\downarrow$      \\
\midrule
\midrule
\multirow{5}{*}{\rotatebox{90}{w cameras}}
\multirow{5}{*}{\rotatebox{90}{single dataset}}
                          & DepthSplat~\cite{xu2025depthsplat}                 & DL3DV 4-view              & 22.30                   & 0.765                   & 0.189                   \\
                          & Ours                                               &                           & \textbf{27.56}          & \textbf{0.869}          & \textbf{0.095}          \\
\cmidrule(lr){2-6}
                          & DepthSplat~\cite{xu2025depthsplat}                 & DL3DV 6-view              & 23.47                   & 0.812                   & 0.154                   \\
                          & Ours                                               &                           & \textbf{29.45}          & \textbf{0.904}          & \textbf{0.068}          \\
\midrule
\midrule
\multirow{7}{*}{\rotatebox{90}{w/o cameras}}
\multirow{7}{*}{\rotatebox{90}{generalizable}}
                          & AnySplat~\cite{jiang25anysplat:}                   & Re10k 2-view              & 17.05                   & 0.626                   & 0.349                   \\
                          & Flare~\cite{zhang25flare:}                         &                           & 23.77                   & 0.801                   & 0.191                   \\
                          & \textcolor{gray}{NopoSplat}~\cite{ye2025noposplat} &                           & \textcolor{gray}{24.06} & \textcolor{gray}{0.820} & \textcolor{gray}{0.178} \\
                          & Ours (v2)                                              &                           & \textbf{25.54}          & \textbf{0.828}          & \textbf{0.158}          \\
\cmidrule(lr){2-6}
                          & AnySplat~\cite{jiang25anysplat:}                   & CO3D 9-view               & 15.87                   & 0.526                   & 0.423                   \\
                          & Ours (v2)                                             &                           & \textbf{22.05}          & \textbf{0.689}          & \textbf{0.365}          \\
\bottomrule
\end{tabular}
}
\caption{\textbf{Feed-forward 3DGS comparison.} Our model outperforms state-of-the-art feed-forward 3DGS models across the board, both with and without input camera poses available, illustrating the strength of \emph{latent} 3D representations.
\textcolor{gray}{NoPoSplat} only trained on RealEstate10k (a potentially unfair comparison), but we include it here for completeness. See~\cref{sec:target_intrinsics} for details of `Ours (v2)'.
}%
\label{tab:vs3dgs}
\end{table}

First, we consider the case where the source views come with known camera poses.
We evaluate our method against DepthSplat, which uses explicit 3D priors in the form of pre-trained monocular depth features, cost volume matching and an explicit 3D representation.
We use their train and test data (DL3DV) and their 4-view and 6-view splits (while taking care to remove source images from evaluation targets) with $256\times256$ center-crops.
Second, we consider the case where the source cameras are unknown and compare against Flare~\cite{zhang25flare:} and AnySplat~\cite{jiang25anysplat:}, which were also trained on a large, diverse dataset.
Following standard practice, we evaluate on 2-view RealEstate10k with the split from~\cite{ye2025noposplat} at $256\times256$.
Like Flare, we use the subset of that split with low- and medium-overlap scenes.
AnySplat uses pre-trained VGGT features, like us.
We further test the latter on the more challenging CO3D~\cite{reizenstein21co3d} $360^\circ$ captures, using the 9-view split from~\cite{wu2024reconfusion} at $512\times512$ resolution.

\paragraph{Results.}

\begin{figure}
\centering
\includegraphics[width=\columnwidth]{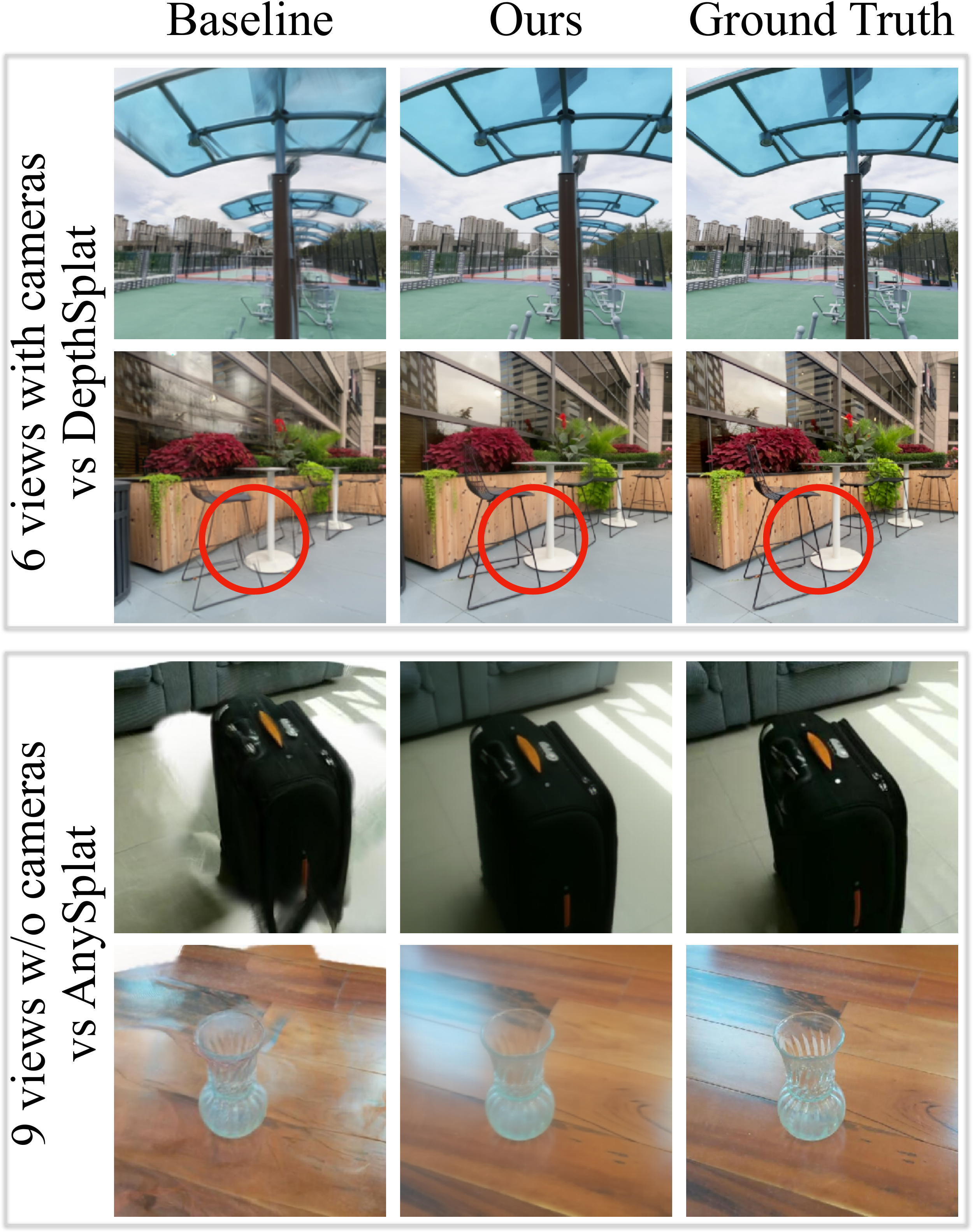}
\caption{\textbf{Comparison to feed-forward 3DGS.} Our model better handles thin parts, reflections, and occlusions, and supports inference with and without source camera poses.}%
\label{fig:vs3dgs}
\end{figure}

As shown in \cref{tab:vs3dgs}, we outperform all methods across the board, often by a significant margin.
\Cref{fig:vs3dgs} shows that our implicit representation better captures reflective surfaces and thin structures.
Note that 3DGS should have no problem representing such elements, but it is evidently difficult to learn a feed-forward model that predicts them accurately.

An even larger difference can be observed in comparison on CO3D in~\cref{tab:vs3dgs}.
\Cref{fig:vs3dgs} shows this can be attributed to the fact that pixel-aligned Gaussian methods completely fail in areas that are not visible in any source view---even with 9 views there are significant occlusions, which are clearly visible (third row of \cref{fig:vs3dgs}).
Occlusion coverage is possible with feed-forward 3DGS~\cite{szymanowicz24splatter,szymanowicz25flash3d}, but AnySplat and Flare force Gaussian-ray alignment---this makes the observed content sharper (notice their competitive LPIPS) but prevents infilling occluded areas.
In contrast, even though our decoder is limited by its deterministic nature, it can, to some extent, `imagine' simple regions, such as the floor (rows 3 and 4 in~\cref{fig:vs3dgs}, more examples in supp.).

\begin{figure}
\centering
\includegraphics[width=\columnwidth]{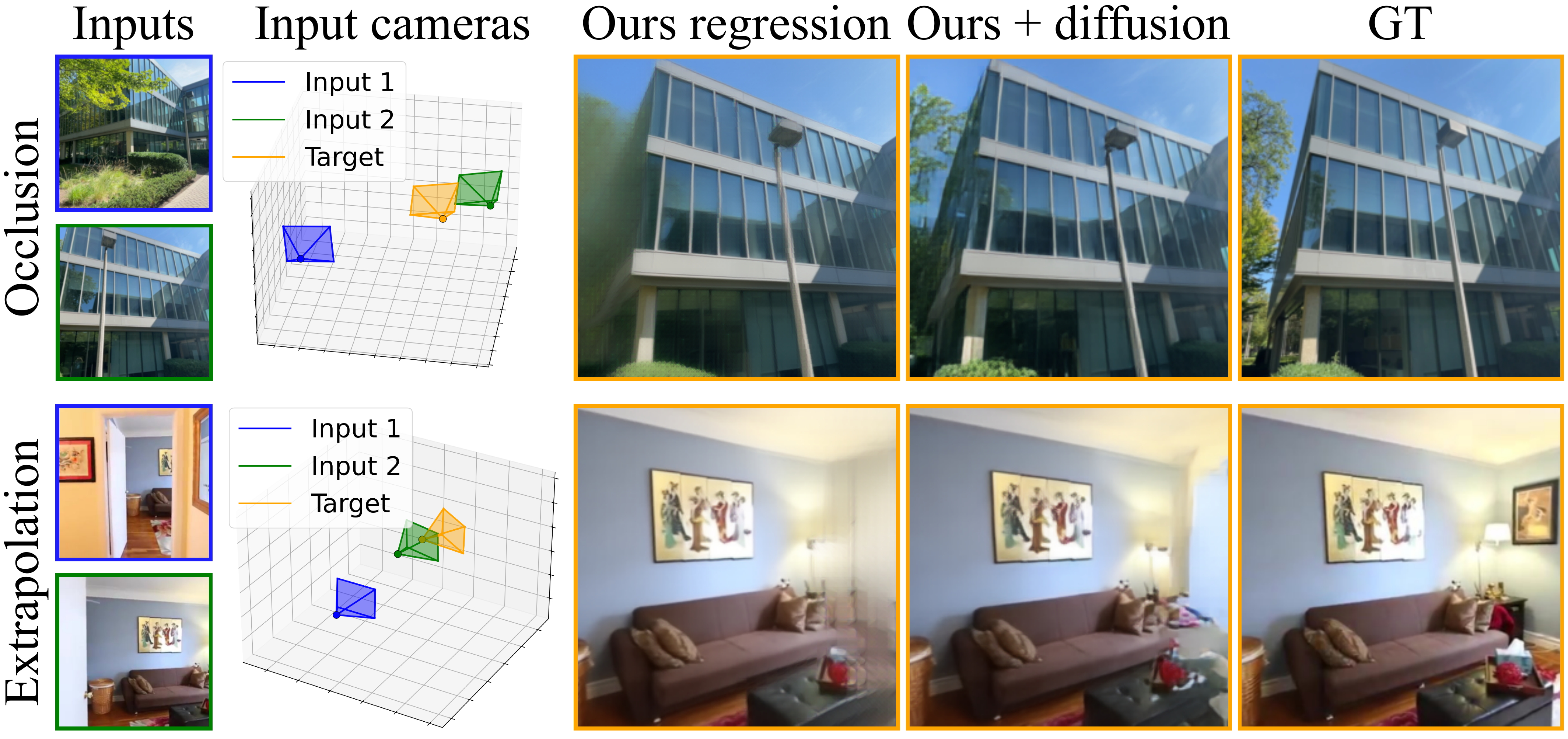}
\caption{\textbf{Diffusion.} Our decoder can be fine-tuned with a diffusion objective, enabling hallucination of unobserved regions.}%
\label{fig:diffusion_regression}
\end{figure}

\subsection{Generative NVS}%
\label{sec:exp-diffusion}

While our model, trained as a deterministic regressor, is capable of simple infilling, high-quality hallucination of unobserved regions requires a generative model.
\method can be adapted to become one, by fine-tuning it to operate as a diffusion model.
In order to do so, we keep the encoder unmodified and frozen, and fine-tune the decoder with a denoising diffusion objective.
The decoder is minimally modified and fine-tuned: we allow the input layer to accept additional channels corresponding to the noisy image, and add adaLN-zero~\cite{peebles23scalable} conditioning on the denoising timestep.

In \cref{fig:diffusion_regression} we demonstrate two scenarios where this formulation is better than the deterministic one:
(1) occlusion and
(2) extrapolation.
In these cases, deterministic models like ours and LVSM regress to the mean and produce blurry outputs, whereas the diffusion variant can hallucinate plausible completions.
Remarkably, our pre-trained model learns diffusion very quickly, after fine-tuning for just 60k iterations, operating in pixel space (without latent diffusion~\cite{rombach22high-resolution}) and using only 12 transformer blocks.

\section{Conclusion}%
\label{sec:conclusion}

We showed the benefits of leveraging strong 3D inductive biases even in NVS models that do not use explicit 3D representations.
We have done so by building our model, \method, from a strong pre-trained 3D reconstruction network.
The resulting model achieves state-of-the-art NVS results when compared to implicit feed-forward NVS models, as well as models that do extract a 3DGS representation of the scene.
Furthermore, our lightweight decoder allows us to render new views at $512\times512$ in real time on a single GPU\@.
Our model is robust, efficient, generalizes well, and requires no camera poses, which makes it practical.
While we have focused on deterministic NVS, experiments show that the model can be applied to generative NVS as well, with further benefits.
We share the code and model checkpoints for the benefit of the open-source community.

\paragraph{Acknowledgments}

We thank Thu Nguyen-Phuoc, Roman Shapovalov, Antoine Toisoul and Zihang Lai for fruitful discussions, David Charatan for assistance with data and Shanghzhan Zhang for confirming evaluation details.
This work was done while S. Szymanowicz and M. Chen were interns at Meta.
S. Szymanowicz is supported by the EPSRC Doctoral Training Partnerships Scholarship
(DTP) EP/R513295/1 and the Oxford-Ashton Scholarship.

{
    \small
    \bibliographystyle{ieeenat_fullname}
    \bibliography{vedaldi_general,vedaldi_specific,main}
}

\appendix
\clearpage
\maketitlesupplementary

\setcounter{figure}{0}
\setcounter{table}{0}
\renewcommand{\thefigure}{A\arabic{figure}}
\renewcommand{\thetable}{A\arabic{table}}

\section{Additional experimental results}

Video results are available on our \href{https://szymanowiczs.github.io/lagernvs}{ project page}.

\subsection{Generalizable, high-resolution  NVS}

Previous works on deterministic Novel View Synthesis typically evaluate quantitative performance in a single-dataset setting at low resolution (256).
To facilitate future comparisons, we include quantitative evaluation on a number of standard benchmarks at $512\times512$, with and without known camera poses in~\cref{tab:high_res_generalizable}.

\begin{table}
\centering
\footnotesize
\resizebox{\columnwidth}{!}{%
\begin{tabular}{llll ccc}
\toprule
Dataset & Views & Posed & Split & PSNR & SSIM & LPIPS \\
\midrule
\midrule
Re10k & 2 & \cmark & {}\cite{charatan24pixelsplat:}
  & 28.99 {\color{gray}(29.05)}
  & 0.900 {\color{gray}(0.901)}
  & 0.149 {\color{gray}(0.147)} \\
Re10k & 2 & \xmark & {}\cite{charatan24pixelsplat:}
  & 27.88 {\color{gray}(28.28)}
  & 0.875 {\color{gray}(0.885)}
  & 0.161 {\color{gray}(0.155)} \\
\midrule
Re10k & 2 & \cmark & {}\cite{zhang25flare:}
  & 26.36 {\color{gray}(26.40)}
  & 0.866 {\color{gray}(0.867)}
  & 0.190 {\color{gray}(0.188)} \\
Re10k & 2 & \xmark & {}\cite{zhang25flare:}
  & 25.11 {\color{gray}(25.64)}
  & 0.833 {\color{gray}(0.848)}
  & 0.210 {\color{gray}(0.201)} \\
\midrule
\midrule
DL3DV & 2 & \cmark & {}\cite{xu2025depthsplat}
  & 21.66 {\color{gray}(21.77)}
  & 0.688 {\color{gray}(0.692)}
  & 0.290 {\color{gray}(0.287)} \\
DL3DV & 2 & \xmark & {}\cite{xu2025depthsplat}
  & 21.27 {\color{gray}(21.33)}
  & 0.666 {\color{gray}(0.670)}
  & 0.303 {\color{gray}(0.301)} \\
\midrule
DL3DV & 4 & \cmark & {}\cite{xu2025depthsplat}
  & 24.90 {\color{gray}(24.94)}
  & 0.778 {\color{gray}(0.780)}
  & 0.189 {\color{gray}(0.188)} \\
DL3DV & 4 & \xmark & {}\cite{xu2025depthsplat}
  & 23.89 {\color{gray}(23.99)}
  & 0.738 {\color{gray}(0.744)}
  & 0.208 {\color{gray}(0.206)} \\
\midrule
DL3DV & 6 & \cmark & {}\cite{xu2025depthsplat}
  & 26.09 {\color{gray}(26.14)}
  & 0.806 {\color{gray}(0.808)}
  & 0.161 {\color{gray}(0.159)} \\
DL3DV & 6 & \xmark & {}\cite{xu2025depthsplat}
  & 24.86 {\color{gray}(24.97)}
  & 0.761 {\color{gray}(0.769)}
  & 0.181 {\color{gray}(0.178)} \\
\midrule
DL3DV & 16 & \cmark & {}\cite{jiang25rayzer:}
  & 25.20 {\color{gray}(25.42)}
  & 0.776 {\color{gray}(0.782)}
  & 0.174 {\color{gray}(0.171)} \\
DL3DV & 16 & \xmark & {}\cite{jiang25rayzer:}
  & 23.31 {\color{gray}(23.49)}
  & 0.713 {\color{gray}(0.719)}
  & 0.214 {\color{gray}(0.211)} \\
\midrule
\midrule
CO3D & 3 & \cmark & {}\cite{wu2024reconfusion}
  & 21.18 {\color{gray}(21.31)}
  & 0.688 {\color{gray}(0.691)}
  & 0.393 {\color{gray}(0.386)} \\
CO3D & 3 & \xmark & {}\cite{wu2024reconfusion}
  & 19.88 {\color{gray}(20.22)}
  & 0.660 {\color{gray}(0.667)}
  & 0.445 {\color{gray}(0.431)} \\
\midrule
CO3D & 6 & \cmark & {}\cite{wu2024reconfusion}
  & 23.41 {\color{gray}(23.65)}
  & 0.728 {\color{gray}(0.733)}
  & 0.326 {\color{gray}(0.317)} \\
CO3D & 6 & \xmark & {}\cite{wu2024reconfusion}
  & 21.37 {\color{gray}(21.65)}
  & 0.680 {\color{gray}(0.684)}
  & 0.388 {\color{gray}(0.377)} \\
\midrule
CO3D & 9 & \cmark & {}\cite{wu2024reconfusion}
  & 24.40 {\color{gray}(24.74)}
  & 0.740 {\color{gray}(0.747)}
  & 0.302 {\color{gray}(0.292)} \\
CO3D & 9 & \xmark & {}\cite{wu2024reconfusion}
  & 22.05 {\color{gray}(22.37)}
  & 0.689 {\color{gray}(0.697)}
  & 0.365 {\color{gray}(0.352)} \\
\midrule
\midrule
Mip360 & 3 & \cmark & {}\cite{wu2024reconfusion}
  & 17.74 {\color{gray}(18.08)}
  & 0.428 {\color{gray}(0.434)}
  & 0.505 {\color{gray}(0.497)} \\
Mip360 & 3 & \xmark & {}\cite{wu2024reconfusion}
  & 17.29 {\color{gray}(17.45)}
  & 0.409 {\color{gray}(0.413)}
  & 0.535 {\color{gray}(0.531)} \\
\midrule
Mip360 & 6 & \cmark & {}\cite{wu2024reconfusion}
  & 19.17 {\color{gray}(19.39)}
  & 0.466 {\color{gray}(0.469)}
  & 0.444 {\color{gray}(0.436)} \\
Mip360 & 6 & \xmark & {}\cite{wu2024reconfusion}
  & 18.81 {\color{gray}(18.97)}
  & 0.446 {\color{gray}(0.447)}
  & 0.472 {\color{gray}(0.466)} \\
\midrule
Mip360 & 9 & \cmark & {}\cite{wu2024reconfusion}
  & 20.19 {\color{gray}(20.39)}
  & 0.487 {\color{gray}(0.493)}
  & 0.412 {\color{gray}(0.402)} \\
Mip360 & 9 & \xmark & {}\cite{wu2024reconfusion}
  & 19.69 {\color{gray}(19.68)}
  & 0.461 {\color{gray}(0.462)}
  & 0.440 {\color{gray}(0.438)} \\
\bottomrule
\end{tabular}
}
\caption{\textbf{Generalizable model performance at $512\times512$.}
To facilitate future comparisons, we report the performance of our generalizable model on a number of standard high-resolution NVS benchmarks, with and without source camera poses. Scores are reported for version 2 of the model in black, described in~\cref{s:method} and~\cref{sec:target_intrinsics}. {\color{gray}Gray} numbers represent scores for the previous version {\color{gray}(v1)}.
}%
\label{tab:high_res_generalizable}
\end{table}

\subsection{Additional comparisons to LVSM}
\begin{figure}
\centering
\includegraphics[width=0.95\columnwidth]{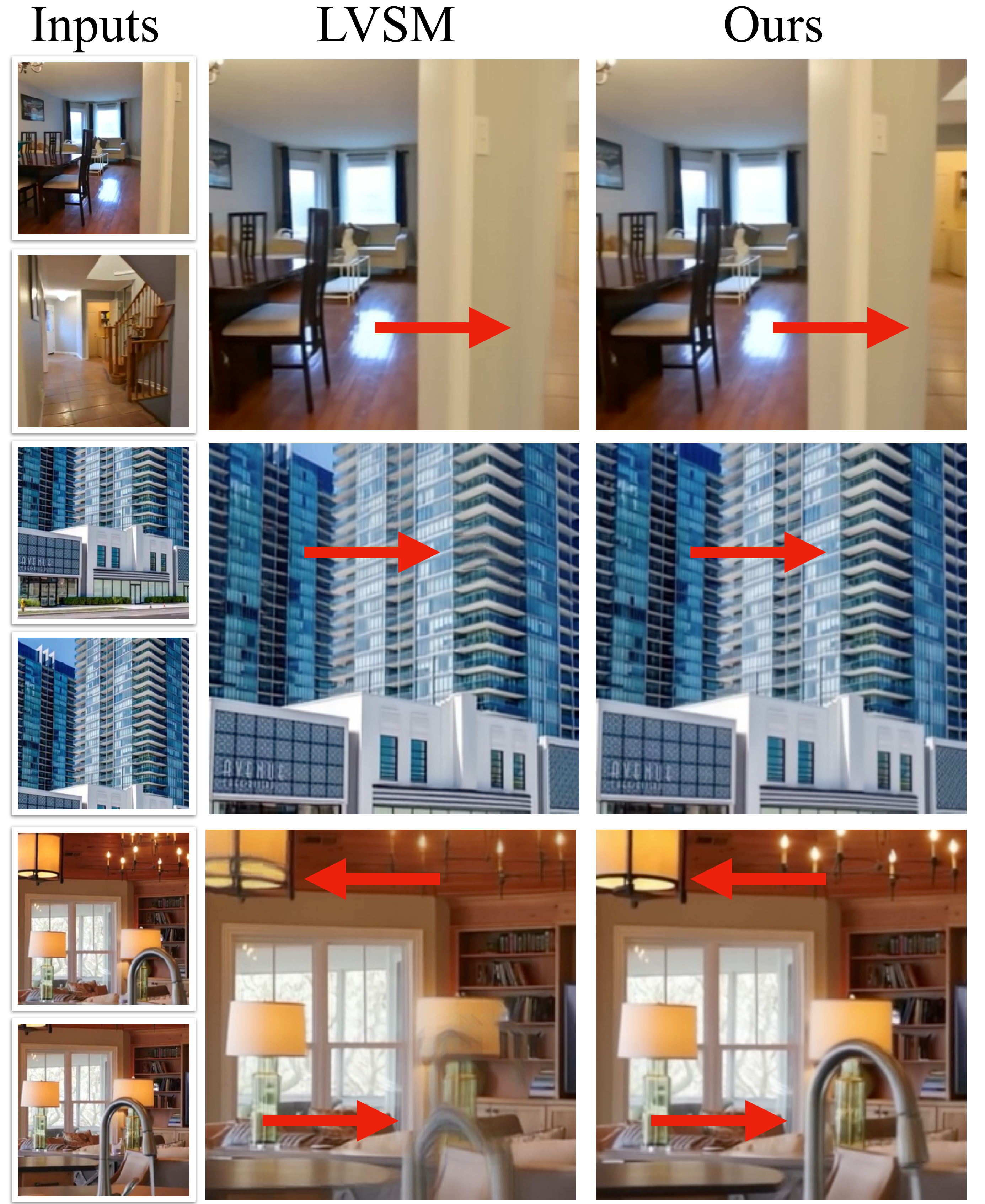}
\caption{\textbf{Additional comparison vs LVSM.} Our model better maintains consistent shape in regions where 2D matching across inputs is challenging (top 2 rows) and shows better monocular depth estimation (bottom row).}%
\label{fig:additional_LVSM}
\end{figure}

Additional qualitative comparisons to LVSM in \cref{fig:additional_LVSM} complement \cref{tab:lvsm_comparison,fig:vs_lvsm_re10k}.
\Cref{fig:additional_LVSM} shows that our model performs better on examples where matching in 2D is difficult, due to low overlap (first row), repeated patterns (second row), and geometry close to the camera (third row).
This indicates that the 3D bias from pre-training helps to disambiguate such challenging cases.

\subsection{SVSM}

SVSM~\cite{kim2026svsm}, a concurrent work, is a variant of LVSM that, similarly to us, adopts an encoder-decoder network with a cross-attention decoder.
Unlike us, however, they focus on efficient compute usage, and use unidirectional cross-attention (\cref{alg:attn_block}).
We instead study the role of 3D pre-training, use a larger encoder and bidirectional cross-attention.
As shown in \cref{tab:svsm}, 
we achieve better quality, although our network is larger.
The training efficiency findings from SVSM are orthogonal to our 3D pre-training and design and can likely be combined for further gains.

\begin{table}[]
\centering
\resizebox{0.5\columnwidth}{!}{%
\begin{tabular}{l ccc}
\toprule
{} & PSNR & SSIM & LPIPS \\
\midrule
SVSM  &  30.01 & 0.910 & 0.096 \\
Ours & \textbf{31.39} & \textbf{0.928} & \textbf{0.078} \\
\bottomrule
\end{tabular}
}
\caption{LagerNVS outperforms SVSM due to greater model capacity and 3D pre-training.}%
\label{tab:svsm}
\end{table}

\begin{figure*}
\centering
\includegraphics[width=0.9\linewidth]{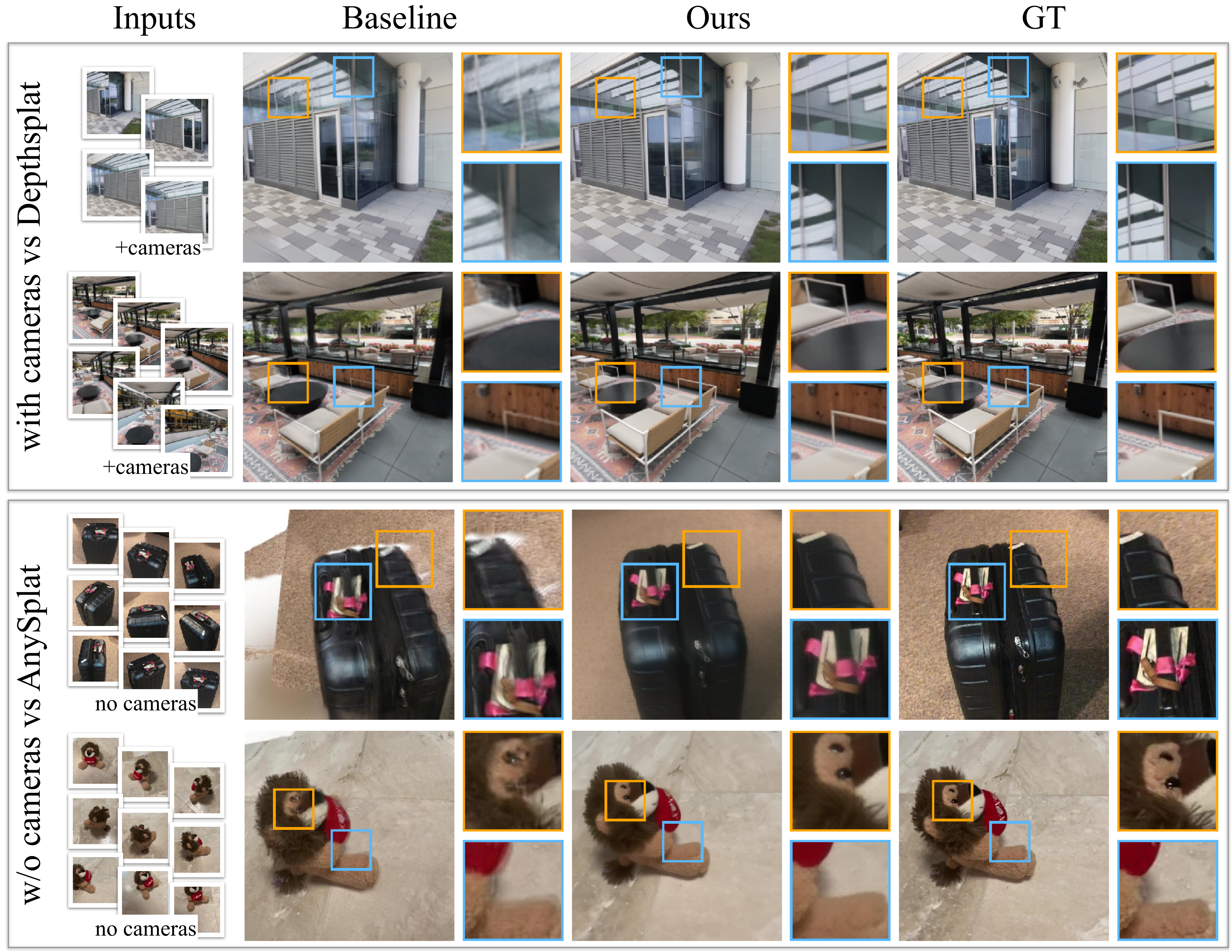}
\caption{\textbf{Qualitative comparison vs Feed-forward 3DGS.} Our model more accurately represents reflective areas (\textcolor{orange}{mirror, top row} and \textcolor{orange}{table, second row}) and thin structures (\textcolor{blue}{metal rod, top row} and \textcolor{blue}{chair arm, second row}).
The global, latent representation leads to better surface alignment (\textcolor{blue}{suitcase handle, third row} and teddy bear, bottom row) and is more robust to occlusions (\textcolor{orange}{floor, third row}).}%
\label{fig:vs3dgs_supp}
\end{figure*}

\subsection{Additional comparisons to 3DGS}

We include additional visual comparisons to feed-forward 3DGS methods in~\cref{fig:vs3dgs_supp}, with and without known cameras, further illustrating that our method improves performance on reflective surfaces, thin structures, and occluded regions.
Other methods can suffer from surface misalignment, especially in $360^{\circ}$ captures.

\subsection{Occlusions}

\begin{figure}
\centering
\includegraphics[width=0.99\columnwidth]{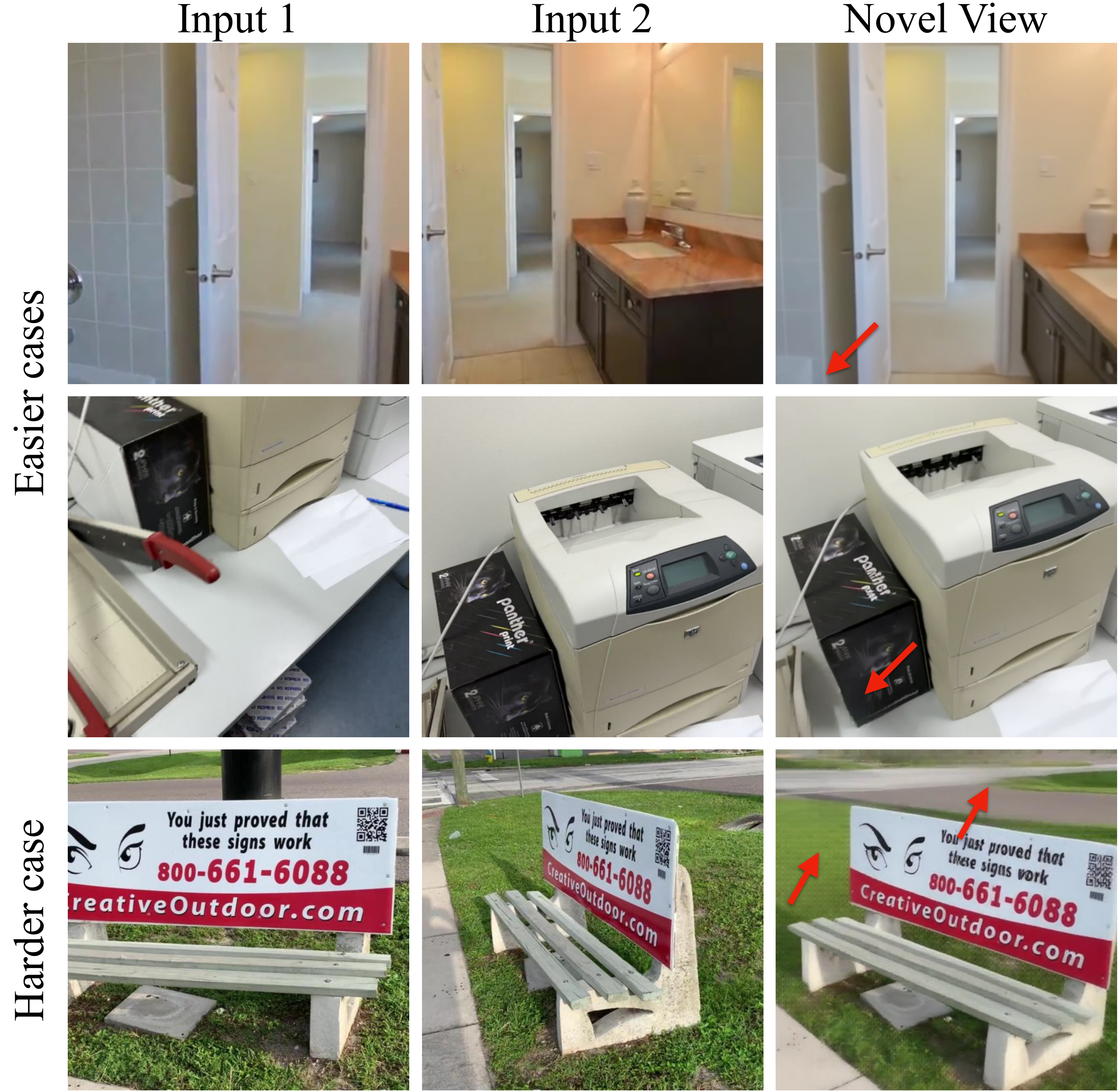}
\caption{\textbf{Occlusions.} 
Our model can handle simple occlusions, such as the corner of the bathtub (top) or the bottom part of the black box (middle).
In more difficult settings (bottom), completions are blurry (as expected), but still reasonable.
}%
\label{fig:occlusions}
\end{figure}

The main model we present is deterministic, so in case of ambiguity (e.g., from occlusions), it tends to regress to the mean.
Interestingly, however, we find that in simple cases, the ``mean'' is a good enough approximation to the ambiguous regions.
In~\cref{fig:occlusions}, we show two examples where the completion is successful: the corner of the bathtub, and the bottom of the box and shelf.
In more difficult examples, as in other deterministic methods (see the limitations section of RayZer~\cite{jiang25rayzer:}), completions are necessarily blurry, but they are still surprisingly reasonable (occluded grass and road in the bottom row of~\cref{fig:occlusions}).
A more principled solution to handling occlusions correctly is to use a \emph{diffusion decoder}, which can sample from the conditional probability distribution.
We illustrated an example of how our model can be fine-tuned as a diffusion model in
~\cref{fig:diffusion_regression} 
of the main paper.
The next section includes more details.

\section{Diffusion}

\paragraph{Implementation details.}
We make two modifications to the architecture of the decoder to allow it to operate as a diffusion model.
First, we add adaLN-zero~\cite{peebles23scalable} conditioning layers to the decoder to condition it on the denoising timestep.
Second, the input patch embedding layer is changed to accept $3$ additional channels, corresponding to the noisy input, leading to a total of $3 + 6=9$ input channels in the decoder.
We employ a DDIM scheduler~\cite{song2020denoising} with 1000 training timesteps, a linear beta schedule from $\beta_{\text{start}}=0.001$ to $\beta_{\text{end}}=0.02$, and train the model to predict the clean sample directly (i.e., $\mathbf{x}_0$ prediction) in pixel space.
We use a zero-SNR noise scheduler~\cite{lin2023common} shifted with a ratio of 4 to account for a relatively high (256$\times$256) resolution~\cite{hoogeboom2023simple}.
Training uses Mean-Squared Error and Perceptual Losses~\cite{johnson16perceptual} with weights $1.0$ and $4.0$, respectively, and a learning rate of $1\times10^{-5}$ for 60k iterations.

\paragraph{Discussion.}
Our diffusion decoder can only generate individual Novel Views, thus generating a video would necessarily result in flicker.
Generating a consistent video would require a video diffusion model as a decoder, or an autoregressive formulation like in~\cite{chan23generative}.
Both are possible, are orthogonal to our contributions, and are an exciting direction for future work.

\begin{figure*}
\centering
\includegraphics[width=0.9\textwidth]{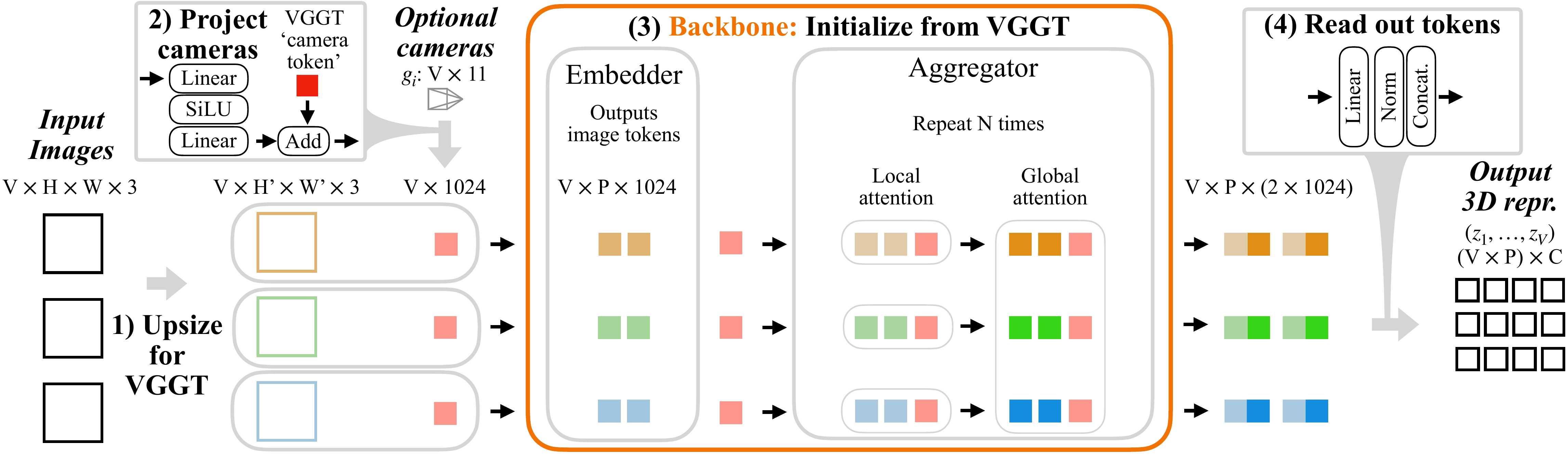}
\caption{\textbf{Encoder.} The encoder takes as source $V$ \textit{images} and, optionally, $V$ \textit{cameras}. The images are upsized (1) to the dimension expected by VGGT, and cameras $\camera$ are projected (2) to conditioning tokens. Image and camera tokens are fed together to the encoder \textcolor{orange}{backbone}, which is initialized with VGGT weights (3). The tokens from the last local attention layer and global attention layer of VGGT are concatenated, after discarding camera tokens. These tokens are then (4) projected to decoder channel dimension $C$ to form the \textit{output latent 3D representation} $\left(\latent_1, \dots, \latent_V \right)$.}%
\label{fig:encoder_detailed}
\end{figure*}

\section{Implementation details}

\subsection{Architecture: encoder}
Here we include additional details of the encoder described in the main paper.
\Cref{fig:encoder_detailed} serves as a visual reference.

Recall that VGGT operates with images with longer side $H' = 518$ or $W'=518$ and token dimension $1024$.
We wish to support images with longer side up to $H = 512$ or $W=512$ (to follow standard resolutions on NVS benchmarks).
Source images are thus first resized to the dimension expected by VGGT~\cite{wang2025vggt}, i.e., with longer side $518$, using bilinear interpolation ((1) in~\cref{fig:encoder_detailed}).

Camera tokens $\camera_i$ are projected to the token dimension $1024$ ((2) in~\cref{fig:encoder_detailed}) using two linear layers with weight size $9\times1024$ and $1024\times1024$, respectively, with SiLU~\cite{elfwing18sigmoid-weighted} activation in between.
The VGGT ``camera token,'' which distinguishes the first camera from the remaining ones, is added to the projected token.
When cameras $\camera_i$ are not available, we set the first 9 elements of the camera vector to 0: $\camera_i[:9] = \boldsymbol{0}$, but keep the scale normalization factor.
Similarly to VGGT, we use the camera pose of the first camera $\camera_1$ as the reference frame.

Image and camera tokens are fed to the encoder backbone ((3) in~\cref{fig:encoder_detailed}), which has weights initialized from VGGT.
The backbone consists of an image embedder (a 24-layer transformer), which operates per-image, and the aggregator (a 48-layer transformer), which interleaves local (per-frame) and global (per-sequence) attention layers.
We keep the architecture of the backbone unmodified; see~\cite{wang2025vggt} for details.
We keep the tokens output by the last local attention layer and the last global attention layer, and concatenate them channel-wise.
The backbone output thus has shape $V \times P \times (2 \times 1024)$, where $P$ is the number of tokens processed by VGGT, after discarding the camera and register tokens.

Next (in (4) in~\cref{fig:encoder_detailed}), a linear layer of weight matrix with size $2048\times 768$ is used to project two features (one from a global attention layer and one from local attention layer) of VGGT channel dimension, $1024$, to the one expected by the decoder, $768$; this is followed by a Layer Normalization~\cite{ba16layer} layer.
Such operations output per-image tokens $\latent_i$, which are concatenated for the whole sequence to form the latent 3D representation $\left(\latent_1, \dots, \latent_V \right)$.

\subsection{Architecture: decoder}
\begin{figure}
\centering
\includegraphics[width=0.6\columnwidth]{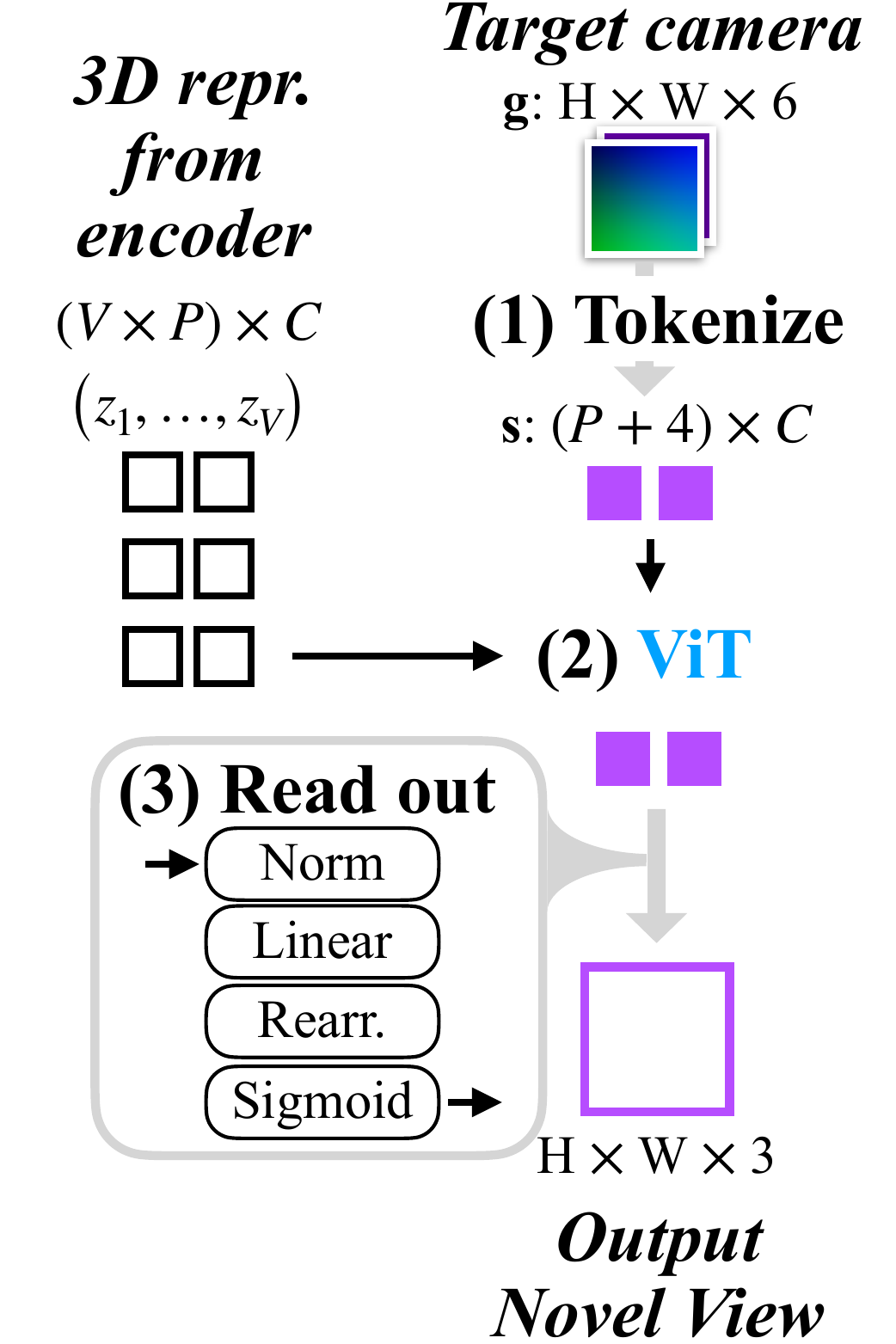}
\caption{\textbf{Decoder.} The decoder tokenizes the \textit{target camera} (1), in the form of a Plucker ray map~\cite{zhang24cameras}. The \textit{scene tokens} (left) and target camera tokens $\targetCameraTokens$ are fed to a Vision Transformer (2)~\cite{dosovitskiy21an-image}, with two variants of the attention mechanism (see text). The register tokens and scene tokens are discarded after the transformer, and \textit{output novel views} are read out from the remaining tokens.}%
\label{fig:decoder-details}
\end{figure}

Here we include additional details of the decoder described in the main paper, with a visual illustration shown in~\cref{fig:decoder-details}.

The decoder receives the \textit{latent 3D representation} of the 3D scene $\left(\latent_1, \dots, \latent_V \right)$, as well as the target camera pose $\camera$.
The camera is first tokenized ((1) in~\cref{fig:decoder-details}) with a strided convolution layer, reshaped, and followed by concatenating 4 register tokens, as described in 
% \cref{sec:decoder} 
Sec 3.2
of the main paper.

Next, the scene representation tokens and the camera tokens $\targetCameraTokens$ are input to the backbone of the decoder, a Vision Transformer~\cite{dosovitskiy21an-image} ((2) in~\cref{fig:decoder-details}).
We use ViT-B, which is a transformer with channel dimension $C=768$, $12$ heads, and $12$ transformer blocks (implemented as in \cref{s:attn_block} and \cref{alg:attn_block}).
The feed-forward layers have a hidden dimension expansion factor of $4$.
Finally, we discard the scene reconstruction tokens and the register tokens, and read out ((3) in~\cref{fig:decoder-details}) the novel view from the updated target camera tokens.
The readout process includes normalization, projecting each of $P$ tokens from channel dimension $C$ to $3 \times r'^2$ (recall $r'$ is the patch size of the decoder), rearranging to the final image dimension $H\times W \times 3$, and a per-element sigmoid activation function.
Following LVSM~\cite{jin25lvsm:}, we do not use biases in the decoder layers.

\subsubsection{Attention mechanism}%
\label{s:attn_block}

As introduced in Sec. 3.2 of the main paper, we consider two variants of the attention mechanism in the transformer blocks (see~\cref{alg:attn_block}).
First, we consider the ``full attention'' variant, where both scene reconstruction tokens and the target camera tokens serve as queries, keys, and values in the attention mechanism.
In the ``cross-attention'' variant, we implement bidirectional cross-attention.
In this variant, both the camera and scene tokens are updated via cross-attention to the other.
The target camera tokens are allowed to self-attend, but there is no self-attention between the scene tokens.
Both sets of tokens are updated via feed-forward layers.
The algorithm for both is described in~\cref{alg:attn_block}.
Such an implementation allows the target camera tokens and latent geometry tokens to communicate with each other and update, without quadratic growth of complexity of the attention operation.

A variant which we do not consider in the main paper, but is commonly used in multimodal transformers such as text-to-image diffusion models~\cite{rombach22high-resolution}, is unidirectional cross-attention.
We detail this variant in~\cref{alg:attn_block} as well, but do not use it in practice, because we find that it performs poorly as the number of source views increases.
Concretely, in~\cref{tab:attn_block}, we find that while the gap in performance to full attention is small ($-0.8$ PSNR) with two source views, it grows substantially (to $-1.9$ PSNR) as the number of views increases to 6.
The unidirectional cross-attention variant is faster because it does not require additional cross-attention and feed-forward update of scene reconstruction tokens.

We select the ``bidirectional cross-attention'' variant for our generalizable model (described in 
Sec 4.2 of the main paper)
as a middle ground between speed and quality.

\begin{algorithm*}
\caption{Transformer block}%
\label{alg:attn_block}
\begin{algorithmic}[1]
\Require Camera tokens $s$, Reconstructed tokens $\mathbf{z}$, Attention Mode $M$

\Procedure{AttentionBlock}{$s, \mathbf{z}, M$}
    \If{$M = \text{``Full attention''}$} 
        \State \Comment{Tokens are concatenated prior to the block: $c = [\mathbf{z} \parallel s]$}
        \State $c \gets c + \text{SelfAttn}\phantom{\text{c}}(Q=c\phantom{\mathbf{z}}, \quad K=c\phantom{\mathbf{z}}, \quad V=c\phantom{\mathbf{z}})$
        \State $c \gets c + \text{FFN}(c)$
        \State \Return $c$ \Comment{Sequence split occurs after all blocks}
        
    \ElsIf{$M = \text{``Unidirectional cross-attn.''}$} 
        \State \Comment{Standard cross-attention block}
        \State $s \gets s + \text{SelfAttn}\phantom{\text{C}}(Q=s\phantom{\mathbf{z}}, \quad K=s\phantom{\mathbf{z}}, \quad V=s\phantom{\mathbf{z}})$
        \State $s \gets s + \text{CrossAttn}(Q=s\phantom{\mathbf{z}}, \quad K=\mathbf{z}\phantom{s}, \quad V=\mathbf{z}\phantom{s})$
        \State $s \gets s + \text{FFN}(s)$
        \State \Return $s$
        
    \ElsIf{$M = \text{``Bidirectional cross-attention''}$} 
        \State \Comment{Bidirectional conditioning}
        \State $s \gets s + \text{SelfAttn}\phantom{\text{C}}(Q=s\phantom{\mathbf{z}}, \quad K=s\phantom{\mathbf{z}}, \quad V=s\phantom{\mathbf{z}})$
        \State $s \gets s + \text{CrossAttn}(Q=s\phantom{\mathbf{z}}, \quad K=\mathbf{z}\phantom{s}, \quad V=\mathbf{z}\phantom{s})$ \Comment{Standard cross-attention}
        \State $\mathbf{z} \gets \mathbf{z} + \text{CrossAttn}(Q=\mathbf{z}\phantom{s}, \quad K=s\phantom{\mathbf{z}}, \quad V=s\phantom{\mathbf{z}})$ \Comment{Cross-attention the `opposite' direction}
        \State $\mathbf{z} \gets \mathbf{z} + \text{FFN}(\mathbf{z})$ \Comment{FFN applied to scene tokens}
        \State $s \gets s + \text{FFN}(s)$ \Comment{FFN applied to camera tokens}
        \State \Return $s, \mathbf{z}$
        
    \EndIf
\EndProcedure
\end{algorithmic}
\end{algorithm*}

\begin{table*}
\centering
\footnotesize
\begin{tabular}{l l c ccc l}
\toprule
{} & {} & Max $\#$ imgs & \multicolumn{3}{c}{Quality (PSNR)} & {} \\
Attention  & Complexity & @ 512 real-time & 2-view & 4-view & 6-view & Params \\
\midrule
\midrule
Full & $\mathcal{O}(V^2)$ & 6 & \textbf{21.30} & \textbf{24.17} & \textbf{24.84} & 85M\\
Unidirectional cross-attn.~(not used)  & $\mathcal{O}(V)$ & \textbf{26} & 20.51 & 22.56 & 22.91 & 113M\\
Bidirectional cross-attn.~(ours) & $\mathcal{O}(V)$ & \underline{9} & \underline{21.02} & \underline{23.65} & \underline{24.54} & 170M \\
\bottomrule
\end{tabular}
\caption{\textbf{Decoder attention variants.} The bidirectional cross-attention mechanism used for the main model offers a good trade-off between real-time rendering speed and NVS quality.
}%
\label{tab:attn_block}
\end{table*}

\subsection{Differences compared to the first version of the model (v1)}%
\label{sec:target_intrinsics}

As noted in \cref{s:method}, the paper has been updated compared to the CVPR version to prevent leaking the camera focal length to the model when no source cameras are passed as input.
The new version, which we call v2, sets the target camera horizontal field of view set to the canonical value $k_0$ when source cameras are unknown.
The value of the canonical horizontal field of view is chosen to be $k_0=53.13^\circ = \text{arctan}(0.5)$ such that horizontal focal length $f_x$ is equal to the image width, $f_x = W$. 
We do so without change to the target image in supervision.
This effectively trains the network to output an image with the same intrinsics as the source images~\cite{bai25recammaster:}.
In the case when source cameras are known, no change is made compared to v1: we set the target camera intrinsics to the same value as the source camera intrinsics.

This change was necessary because during training the source and target images have the same focal length.
As a consequence, specifying the focal length for the target images could potentially ``leak'' the focal length information of the source cameras.
Moreover, specifying target camera intrinsics different from (potentially unknown) source images resulted in deteriorated performance in model v1 due to a train-test mismatch.
Providing a nominal value $k_0$ as horizontal field of view solves both issues.
First, it provides no information about source camera intrinsics.
Second, such nominal value can be applied both at training and testing time, thus removing the train-test mismatch.

After the change, the new model has slightly worse performance on the benchmarks (but is still state-of-the-art by a good margin).
Table~\cref{tab:high_res_generalizable} shows the difference.
We have updated the results in the paper to refer to `v2' when relevant, and marked this explicitly in~\cref{tab:vs3dgs}.
Note that only the `generalizable' variant of the model is affected by this change (because that is the only model we train with unposed source images), so some of the rows are unaffected.

Note also that model v1 is completely valid for the case where source cameras are passed as input, but we prefer for the users to use model v2 as default for all cases as the performance difference is minor.

\subsection{Training details}%
\label{sec:training-details}

\paragraph{Data}
One advantage of building on multi-view stereo models like VGGT~\cite{wang2025vggt} is that they generalize well due to training on a large collection of datasets.
In line with this work, we consider a rich mix of datasets to supervise our model too, including
TartanAir~\cite{wang2020tartanair},
Eden~\cite{le21wacv},
ARKit~\cite{dehghan2021arkitscenes},
BlendMVS~\cite{yao2020blendedmvs},
Hypersim~\cite{roberts21hypersim:},
UCo3D~\cite{liu2025uco3d},
Taskonomy~\cite{zamir18taskonomy:},
RealEstate10k~\cite{zhou18stereo},
StaticThings~\cite{schroeppel2022robust},
NeSF~\cite{vora2021nesf},
MVSSync~\cite{huang18deepmvs:},
WildRGBD~\cite{xia24rgbd},
and approximately 45k Internet-style videos, whose cameras were annotated using a structure-from-motion pipeline~\cite{schonberger16structure-from-motion} with post-filtering.
We sample the datasets with variable proportions, balancing their relative size and diversity.

\paragraph{Shared hyperparameters}
Throughout all experiments, we use AdamW~\cite{kingma15adam} optimizer with beta coefficients $(\beta_1, \beta_2) = (0.9, 0.95)$ and weight decay of $0.05$.
We skip iterations when the gradient norm is larger than $5.0$.
The learning rate is warmed up linearly for $3k$ iterations, followed by cosine annealing to zero with warm restarts.
We detail the remaining hyperparameters in~\cref{tab:hyperparams}.

\begin{table*}[t]
\centering
\footnotesize
\resizebox{0.7\textwidth}{!}{%
\begin{tabular}{ll cccccc}
\toprule
Experiment &  Iter. & LR & Batch & Data & Source Views & Res & Grad clip \\
\midrule
LVSM comparison Tab 1. (a-c) & 100k & $4e-4$ & $64$ & Re10k & 2 & 256 & 1.0\\
LVSM comparison Tab 1. (d-f) & 100k & $4e-4$ & $512$ & Re10k & 2 & 256 & 1.0\\
\midrule
Ablations Tab 2. & 100k & $1e-4$ & $64$      & All & 2--4 & 256 & 3.0\\
\midrule
3DGS Tab 3. DL3DV posed & 100k & $4e-4$ & $512$ & DL3DV & 2--6 & 256 & 3.0\\
\midrule
Final model, 3DGS Tab 3. Unposed and Tab 4. & 250k & $1e-4$ & $512$ & All & 1--10 & 512 & 3.0 \\
\bottomrule
\end{tabular}
}
\caption{\textbf{Hyperparameters.} We detail the training hyperparameters for each experiment.
The model used for comparison in ``3DGS Tab 3. Unposed'' can also accept camera poses as conditioning, and we use it as our final model.}%
\label{tab:hyperparams}
\end{table*}

\paragraph{Multiple targets per one scene encoding.}
The forward pass through the encoder, which has to process $\numViews\geq 1$ images, is much slower than a pass through the decoder, which processes a single target view.
We thus rebalance the costs by predicting $\numTargetViews \geq 1$ target views for each set of source images in a single batch (these images are rendered in parallel but still independently).

\paragraph{View, image, and camera augmentation.}

We randomly sample the number of source views $\numViews$ during training for each batch to allow our model to accept a variable number of source images at test time.
We randomly use $1-10$ source views in each batch, inclusive.
We sample (uniformly) an even number of source views with probability $4:1$ to sampling an odd number of views.
Each example has a fixed number of target views $V_{target}=8$.
We randomly drop the camera pose token for $40\%$ of training examples, jointly training for posed and unposed NVS\@.
Aspect ratio is sampled uniformly in the log domain between $[0.5, 2.0]$.

\paragraph{Scene scale.}

To generate the target image $I$, an NVS algorithm is given the target camera $\camera$, which specifies how much the camera moves with respect to the imaged scene.
Due to scale ambiguity, cameras and 3D scene are generally only defined up to a common scaling factor.
The NVS algorithm must determine the relationship between the scene scale and the target camera scale to interpret the target viewpoint correctly.

When there are several source views with know source cameras, the NVS algorithm can infer the  scale of the scene relative to those cameras from triangulation, and use this information to interpret the scale of the target camera too.
However, when the source cameras are \emph{not} specified as input, or when there is a single source view and triangulation is not possible, the NVS algorithm requires additional information and/or assumptions to find the relation between the scene scale and the target camera scale.

A possible approach is to assume that the target camera motion is expressed meters, and that the NVS model can implicitly perform metric reconstruction of the scene.
We discount this approach because it is challenging to collect metric data at large enough volumes for training accurate metric NVS systems\@.

Another approach is to define the scale to be the maximum distance of any camera pose from the reference (first) camera, i.e.,  $w_1 = \text{max}_i(\|t_i\|_{2})$.
This is convenient as camera poses can usually be estimated reliably and quickly  (e.g., from the 5-point algorithm~\cite{triggs99camera}).
Camera poses are also available for the majority of evaluation benchmarks, thus making this normalization method well-suited for quantitative evaluations.

On the other hand, when there is a single source image, or when all the source cameras overlap, $w_1=0$ and using this scale is not possible.
We thus consider a second method of normalization, based on the scale of the points observed in the scene, which we define as the average distance from the reference camera to all points visible in all source images 
\begin{equation*}
    w_2 = \frac{1}{\sum_i \sum_u \sum_v  \mathbbm{1}_{iuv}} \sum_i \sum_u \sum_v \mathbbm{1}_{iuv} \|x_{iuv}\|_2,
\end{equation*}
where $\mathbbm{1}_{iuv}$ is the indicator function that evaluates to 1 if pixel in view $i$ at coordinate $u,v$ holds a valid depth, and $x_{iuv}$ is the world-coordinate location of the point.

We train our model to use either normalization method, depending on what is available at inference time.
We do this by inputting both normalization factors $w_1$ and $w_2$ to the network as part of the camera parameters; in other words, highlighting the target camera parameters, the NVS function is of the type:
$$
I = f(\cameraRotation,\cameraTranslation,\cameraIntrinsics, w_1,w_2, \dots)
$$
For each given training scene and camera, this gives us a certain value for the tuple
$
(\cameraRotation,\cameraTranslation,\cameraIntrinsics,
w_1, w_2)
$.
Furthermore, we can rescale the scene and cameras by a common factor to obtain a different tuple
$
(\cameraRotation,\lambda \cameraTranslation,\cameraIntrinsics,
\lambda w_1, \lambda w_2)
$, obtaining another valid training example.
Because one scaling factor may be unavailable at inference time, at training time we also randomly drop either of them out, setting it to zero, in which case the model learns to rely only on the other.

In practice, this is done as follows.
\begin{enumerate}
\item If only camera poses are available for a particular training scene, then $w_2=0$ and we choose $\lambda$ so that that $\lambda w_1 := 1/1.35$.

\item If both camera poses and points are available, the we proceed as follows.
\begin{itemize}
\item We choose a $\lambda$. To do so, with 50\% probability, we choose $\lambda$ so that $\lambda w_1 = 1/1.35$ (following LVSM) and with 50\% probability so that $\lambda w_2 = 1$.

\item We choose which, if any, scaling factor to drop out.
With 1/3 probability, we drop out $w_1$, $w_2$ or neither, thus training with 
$(0, \lambda w_2)$,
$(\lambda w_1, 0)$, or
$(\lambda w_1, \lambda w_2)$.
\end{itemize}
\end{enumerate}

At test time, whenever possible, we use the scaling factor $\lambda w_1:=1/1.35$ and pass $w_2=0$ to the network.
As discussed above, this is a viable option whenever there are at least two cameras with non-overlapping origins.
During quantitative evaluation, we compute $\lambda$ in the dataloader so that $\lambda w_1:=1/1.35$ as required.
If the evaluation is on unposed images, as discussed in the main paper, we set $\camera_i$ to the null token, keeping only scale parameter $\lambda w_1$, so that the source cameras are not available to the model, but scale is unambiguous.
The network is able to implicitly learn the meaning of this scaling protocol during training, and thus uniquely estimate the image $\image$ for the target camera $\camera$.

When such scaling is not possible due to single source image or overlapping camera origins, we use the other scaling factor.
In the dataloader, we can compute $\lambda$ so that $\lambda w_2 = 1.0$, and pass
$
(\lambda w_1, \lambda w_2)
=
(0, 1.0).
$
In some cases, we may not be able to compute $w_2$ either (due to a lack of depth information).
Here, we simply pass $(w_1,w_2)=(0,1)$, which means that the model implicitly understand the target camera translation to be expressed as a fraction of the scale of the points it observes in the scene.

Such training protocol was necessary for supporting single-view NVS with our main generalizable model.
When training only with $w_1$ (scaling based on cameras) and evaluation with one source image, the model was capable of rendering images under rotation around center of projection of the source camera, but unable to render images under camera translation.
Adding scaling based on points during training, using $w_2$, resulted in successful renders under camera translation (as shown in 
\cref{fig:single_view} 
in the main paper).

\section{Limitations}

Our model is trained with identical camera intrinsics for all source images, equal to target camera intrinsics.
As a consequence, providing source images with different focal lengths or different target camera intrinsics can lead to deteriorated performance.
A more general version of the model should be trained to respond correctly to an arbitrary target focal length and relax the constraint that focal lengths are the same within the source cameras and with the target camera.
This could be achieved by focal length randomization during training.

Next, the model is not capable of high-quality hallucination of unseen regions, and exhibits blurry renderings with block artifacts when rendering unobserved regions.
Occasionally, when rendering a video, such block artifacts result in an impression of flicker, which often stems from uncertainty in geometry estimation, unobserved regions, or difficulty in estimating source camera poses.
Additionally, regions with high-frequency patterns, such as grass or trees, are systematically poorly represented by our model---investigating this failure mode is an interesting avenue for future work.
Moreover, our model is suited only to static data, did not include humans in the training data, and did not include images with distortion (e.g., fish-eye).
As a consequence, we do not expect the model to work well in such scenarios.

\end{document}